%% file: main.tex
\documentclass[11pt]{article}

\usepackage[final]{acl}

\usepackage{times}
\usepackage{latexsym}

\usepackage[T1]{fontenc}
\usepackage[utf8]{inputenc}
\usepackage{microtype}

\usepackage{amsmath}
\usepackage{amssymb}
\usepackage{graphicx}
\usepackage{float}
\usepackage{booktabs}
\usepackage{enumitem}
\usepackage{placeins}
\usepackage{multicol}
\usepackage{fvextra}
\usepackage[breakable]{tcolorbox}
\usepackage{etoolbox}
\usepackage{tabularx}
\usepackage{fontawesome5}
\usepackage{pgfplots}
\pgfplotsset{compat=1.16}
\usepgfplotslibrary{groupplots}
\usetikzlibrary{patterns,shadows,backgrounds}

\definecolor{clrHRM}{HTML}{F9BC60}
\definecolor{clrSIC}{HTML}{2A687A}
\definecolor{clrTIS}{HTML}{E8985E}
\definecolor{clrMHRD}{HTML}{5B8C85}
\definecolor{clrBOX}{HTML}{B8B8B8}
\definecolor{clrHRMlt}{HTML}{FCDEA8}
\definecolor{clrSIClt}{HTML}{7FAAB5}
\definecolor{clrTISlt}{HTML}{F2C9A8}
\definecolor{clrMHRDlt}{HTML}{A3C5C0}
\definecolor{clrBOXlt}{HTML}{DCDCDC}
\definecolor{clrParse}{HTML}{E8985E}
\definecolor{clrRuntime}{HTML}{F9BC60}
\definecolor{clrWrong}{HTML}{2A687A}
\definecolor{clrSolved}{HTML}{5B8C85}

\newcommand{\codeurl}{https://github.com/aditya-pola/ISA-Bench}
\newcommand{\projecturl}{https://aditya-pola.github.io/ISA-Bench/}
\definecolor{btnbg}{HTML}{ffffff}
\definecolor{btnborder}{HTML}{000000}
\newcommand{\linkbutton}[3]{%
  \href{#1}{\tikz[baseline=(text.base)]{\node[fill=btnbg, draw=btnborder, rounded corners=4pt,
    inner xsep=8pt, inner ysep=4pt] (text) {\textcolor{black}{#2}\kern0.4em\textcolor{black}{\small #3}};}}}

\title{ISA-Bench: A Benchmark for Computational Reasoning Across Instruction Set Architectures}
\author{
  \textbf{Aditya Pola\textsuperscript{1}},
  \textbf{Arkaprava Majumdar\textsuperscript{1}},
  \textbf{Vineeth N. Balasubramanian\textsuperscript{1,2}}
\\
  \textsuperscript{1}Indian Institute of Technology Hyderabad,
  \textsuperscript{2}Microsoft Research India
\\[8pt]
  \linkbutton{\projecturl}{\faGlobe}{Project Page}\hspace{1em}%
  \linkbutton{\codeurl}{\faGithub}{Code}
}

\begin{document}
\maketitle

\begin{abstract}
Large language model code generation benchmarks primarily evaluate well-resourced languages like Python and Java, where models benefit from abundant training data. They provide limited evidence about reasoning in unfamiliar computational models: deriving arithmetic from a single subtract instruction, coordinating parallel programs across communicating nodes, or wiring logic gates into circuits. We present ISA-Bench, a benchmark of programming games with constrained instruction sets. For each game we provide a full execution stack (parser, VM, and verifier), enabling automated evaluation with structured feedback for iterative refinement. Reasoning models achieve higher average solve rates than code-specialized and general-purpose models, but unfamiliar syntax remains a major source of failure. Models solve more tasks with iterative feedback, though the gains vary substantially across architectures. We introduce a reasoning--execution gap (REG) analysis that reveals a recurring disconnect between identifying a plausible computational strategy and expressing it as a correct program in the target ISA.
\end{abstract}

\section{Introduction}

Large language model code generation benchmarks have scaled from function-level completion to repository-level software engineering and competitive programming. Their target languages, predominantly Python, Java, and C++, are abundantly represented in pretraining corpora. When the target language is less represented in training data, code generation accuracy degrades, and drops further for domain-specific languages with limited web presence. At the far end of this spectrum lie the custom instruction sets of assembly-style programming games, languages unlikely to appear at scale in web-crawled training data, where a model must reason about the target language from its specification alone. This end of the spectrum remains largely unbenchmarked.

Programming puzzle games built around constrained instruction sets offer controlled environments for studying code generation under such conditions. Each game provides an instruction set specification, a set of tasks with deterministic test cases, and a verifier that confirms correctness through execution. Because these games were originally designed to teach computational thinking, they come with built-in difficulty progressions and self-contained specifications, making them natural benchmarks. The five games we select require models to write programs in unfamiliar instruction sets, from writing arithmetic using a single subtract-and-branch operation, to coordinating parallel programs across a grid of communicating nodes, to wiring logic gates into circuits. Together, they cover a range of computational models within a shared evaluation structure.

Solving tasks in these instruction sets requires reasoning about the computational model itself rather than mapping natural-language intent to familiar syntax. The model must infer how to construct operations that the instruction set does not provide, plan resource usage under architectural constraints, and in some cases coordinate multiple programs that execute concurrently. ISA-Bench tests program generation under these constraints, although prior exposure to game solutions or reusable programming patterns cannot be ruled out.

Our evaluation of 14 open-weight models (8B to 32B parameters) surfaces several patterns. Parse errors are the largest first-attempt outcome category in each game's reported distribution, exceeding 60\% on HRM, SIC-1, and BOX-256. Thinking models have higher average solve rates than code-specialized and general-purpose models, but these comparisons do not control for differences in scale, training, or inference behavior. A separate reasoning--execution gap analysis records cases where plausible plans receive lower implementation scores. Its game-level differences provide a diagnostic perspective rather than a causal explanation of the solve rates.

Our contributions are:
\begin{enumerate}
    \item \textbf{ISA-Bench}: five programming game environments with parsers, execution engines, and test-case verifiers for each game's instruction set, covering 92 tasks across distinct computational reasoning primitives.
    \item A baseline evaluation of 14 open-weight models on program generation in instruction sets minimally represented in pretraining data, under single-shot and iterative settings.
    \item A failure-mode taxonomy and reasoning--execution gap (REG) analysis distinguishing planning scores from implementation scores, alongside a descriptive analysis of iterative retry behavior.
\end{enumerate}

\input{figures/fig_overview}
\input{figures/evaluation_pipeline}

\section{Related Work}

Existing code-generation benchmarks predominantly target mainstream programming languages. Related work also examines domain-specific language generation through grammar prompting \citep{wang2023grammar} and hardware generation through Verilog/RTL benchmarks such as VerilogEval and RTLLM \citep{liu2023verilogeval,lu2024rtllm}. ISA-Bench complements these efforts by evaluating multiple constrained computational models within a common execution framework and introducing a reasoning--execution gap (REG) analysis of generated programs. Table~\ref{tab:related_work_benchmarks} summarizes key differences.

\begin{table}[t]
\centering
\small
\renewcommand{\arraystretch}{1.15}

\begin{tabularx}{\columnwidth}{@{}l r l X@{}}
\toprule
\textbf{Benchmark} & \textbf{Tasks} & \textbf{Lang.} & \textbf{What it tests} \\
\midrule
HumanEval         & 164  & Python     & Function completion, pass@$k$ \\
MBPP               & 974  & Python     & Entry-level synthesis \\
CodeContests       & 13k  & C++/Py     & Algorithmic reasoning \\
SWE-bench          & 2.3k & Python     & Repository issue resolution \\
LiveCodeBench      & Live & Python     & Contamination-free, self-repair \\
\midrule
\textbf{ISA-Bench} & \textbf{92} & \textbf{5 ISAs} & \textbf{Unfamiliar ISA synthesis, reasoning--execution gap} \\
\bottomrule
\end{tabularx}
\caption{Comparison with code-generation benchmarks. Existing work predominantly targets mainstream programming languages. ISA-Bench evaluates synthesis in unfamiliar instruction sets.}
\label{tab:related_work_benchmarks}
\end{table}

The standard evaluation paradigm for LLM code generation centers on functional correctness measured through hidden unit tests. HumanEval introduced the pass@$k$ estimator for this purpose \citep{chen2021humaneval}, and MBPP extended the approach to 974 entry-level Python tasks \citep{austin2021mbpp}. Subsequent benchmarks have raised the difficulty ceiling. APPS and CodeContests target full-program synthesis with contest-style algorithmic reasoning \citep{hendrycks2021measuringcodingchallengecompetence,li2022alphacode}, LiveCodeBench provides contamination-resistant, continuously refreshed problems with explicit self-repair evaluation \citep{jain2024livecodebench}, and SWE-bench tests repository-level issue resolution on real Python projects \citep{jimenez2024swebench}. These benchmarks are foundational, but they operate exclusively in mainstream languages and toolchains, and therefore do not isolate whether models can infer computation in an unfamiliar ISA from its specification alone.

Related work in program synthesis and executable datasets has likewise emphasized semantic evaluation. ExeBench pairs millions of compilable C functions with input--output examples, enabling evaluation against real execution rather than surface-form similarity \citep{ExeBench}. Jigsaw augments LLM generation with program-analysis and synthesis-based post-processing, demonstrating that symbolic constraints can improve end-to-end accuracy in API-oriented synthesis \citep{jain2021jigsawlargelanguagemodels}. ISA-Bench differs from these efforts by evaluating direct generation of executable programs in low-resource, assembly-like languages with task-specific virtual machines and verifiers, rather than operating in familiar C or Python environments or relying on post-hoc repair.

At the low-level end of the spectrum, assembly-like program generation has appeared primarily in adversarial evolution and translation settings rather than as benchmarked text-to-program synthesis. Prior work has evolved assembly programs for the adversarial CodeGuru environment \citep{maliukov2024assembly} and studied LLM-driven self-play in Core War \citep{kumar2026drq}. Neural compilation has been explored as translation from a semantics-bearing source, including C$\rightarrow$x86 compilation \citep{zhang2024cx86} and CISC$\rightarrow$RISC assembly transpilation \citep{armengol2024cisc}. In contrast, ISA-Bench requires the model to derive both the algorithm and its executable implementation from a natural-language task description and an ISA reference, without a source program to translate.

Iterative refinement through execution feedback has received growing attention. The pass@$k$ framework measures search over candidate programs through independent sampling \citep{chen2021humaneval}, and self-debugging approaches have demonstrated that execution feedback can improve sample efficiency \citep{chen2024selfdebug}, though self-repair gains are often modest and highly uneven once repair costs are accounted for \citep{olausson2024selfrepair}. ISA-Bench evaluates iterative refinement across all five environments and additionally separates failures of computational planning from failures of low-level implementation. This distinction is especially informative when comparing performance across multiple unfamiliar instruction sets, where the same solve rate can mask qualitatively different failure modes.

\section{ISA-Bench}

ISA-Bench comprises five programming game environments. Each game follows the same evaluation structure. The model receives a task description and an instruction set reference (the opcodes, operand formats, and execution rules that define the game's language), writes a program in that instruction set, and submits it to the game's execution environment, which runs the program and checks it against expected outputs. What varies across games is the instruction set, the execution model, and the kind of reasoning the tasks demand. Figure~\ref{fig:overview} illustrates the five environments.

For each game, we implement three components:
\begin{enumerate}[nosep,leftmargin=*]
\item A \textbf{parser} (or assembler) reads generated source text and translates it into the game's internal representation, whether a memory image, an instruction list, or a circuit graph. Syntax errors and invalid opcodes are caught at this stage.
\item An \textbf{execution engine} runs the parsed program according to the game's computational model, stepping through instructions, cycling nodes, or propagating signals. Runtime errors such as infinite loops or illegal memory access surface here.
\item A \textbf{verifier} runs the program against the task's test cases and compares outputs to expected results. This determines correctness.
\end{enumerate}
\noindent Together, these components accept arbitrary generated code and return structured feedback, enabling both single-shot evaluation and iterative refinement. Figure~\ref{fig:pipeline} illustrates this pipeline.

\input{figures/fig_solverate}

\paragraph{SIC-1 (Single Instruction Computer)~\cite{krinke2022sic1}.}
Tests whether a model can construct complex behavior from a single primitive operation. The instruction set has one opcode (SUBLEQ), which takes three operands: it subtracts one memory value from another and branches if the result is non-positive. There are no opcodes for arithmetic, logic, or control flow; every operation must be derived from subtract-and-branch over 256 bytes of linear memory. The architecture is Turing-complete~\cite{mavaddat1988urisc}. The 31 tasks range from echoing input to addition, sorting, and self-hosting interpreters (full list in Appendix~\ref{app:tasks}).

\paragraph{Human Resource Machine~\cite{tomorrowcorp2015hrm}.}
Requires the model to plan a sequence of individual operations to transform an input stream into an output stream. The instruction set has 11 opcodes (load, store, add, subtract, conditional branch, and I/O) with direct and indirect addressing. The model operates through a single register and an indexed memory space, so to perform any computation, values must be loaded, manipulated, and stored one at a time. The 36 tasks range from copying values between queues to computing Fibonacci sequences, sorting lists, and traversing indirect-address chains.

\paragraph{TIS-100~\cite{zachtronics2015tis100}.}
Asks the model to solve a task by writing separate programs for multiple nodes that run concurrently and pass values between each other. The architecture is a $4\times3$ grid of 12 nodes, each executing its own program using 13 opcodes (arithmetic, moves, and conditional jumps). The model outputs programs for individual nodes, marked by identifiers (\texttt{@0}, \texttt{@1}, \texttt{@2}, \ldots). Nodes exchange values through directional ports, where each transfer blocks until the receiving node reads the value. Input streams feed into the top of the grid, outputs are collected from the bottom. The model must decide which nodes to program, what each one computes, and how values flow between them. The 7 tasks include doubling each value in a stream, reversing a sequence, and multiplying pairs of inputs.

\paragraph{MHRD~\cite{funghisoft2017mhrd}.}
Evaluates whether a model can express boolean functions as circuit structure. The model writes a hardware description by declaring input and output pins, instantiating components, and wiring them together. The only primitive is a NAND gate, so all circuits must be composed from it. Correctness is determined by evaluating the circuit against a complete truth table. The 8 tasks progress from NOT and AND gates to multiplexers, half-adders, and full-adders~\cite{nisan2021nand2tetris}.

\paragraph{BOX-256~\cite{kiili2016box256}.}
The model must write a short assembly program that produces a target pixel pattern on a $16\times16$ grid. The grid's 256 cells are the only memory: the program's instructions are loaded into the first cells, and the program executes by reading and writing to the same grid. Every cell is simultaneously rendered as a colored pixel, so the program's own instructions are visible in the image. Drawing a pixel means writing a value to memory. The 13 opcodes cover arithmetic, control flow, and setting the color of a cell at a given coordinate. Programs run for a fixed number of cycles, after which the grid's pixel state is compared to the target image. The 10 tasks specify target patterns from rectangles and checkerboards to spirals and fractals.

\medskip
\noindent In total, ISA-Bench covers 92 tasks across the five environments, with difficulty progressions inherited from each game's design (Appendix~\ref{app:isa} summarizes each ISA; Appendix~\ref{app:examples} provides worked examples).

\section{Experiments}

\subsection{Models}

We evaluate 14 open-weight language models ranging from 8B to 32B parameters. Thinking models (Qwen3 8B/14B/32B~\cite{qwen3}, Gemma-4 31B~\cite{gemma4}, DeepSeek-R1 32B~\cite{deepseekr1}, Phi-4-Reasoning~\cite{abdin2025phi4reasoning}) generate extended reasoning traces before producing code. Code-specialized models (Qwen3-Coder 30B~\cite{qwen3}, Qwen2.5-Coder 14B/32B~\cite{hui2024qwen25coder}, Codestral 22B~\cite{codestral2024}, DeepSeek-Coder-V2 16B~\cite{deepseekcoder2}, Devstral-Small-2~\cite{devstral2025}) are specialized for code generation or software-engineering tasks. General-purpose models (Granite-4.1 8B~\cite{granite412026}, Phi-4 14B~\cite{abdin2024phi4}) are designed for broad instruction-following tasks.

We restrict evaluation to open-weight models because ISA-Bench is designed as a base benchmark that the community can extend through fine-tuning, reinforcement learning, and interpretability studies. Open-weight models ensure full reproducibility (model weights, hyperparameters, and prompts are all publicly available) and allow researchers to build on these baselines by analyzing internal representations or training on ISA-specific data. Closed-source models, by contrast, expose only limited, edited chain-of-thought traces and final outputs, restricting downstream analysis to behavioral observations alone.

\input{figures/fig_failure}

\subsection{Evaluation Protocol}
\label{sec:eval_protocol}

Each model receives a system prompt containing the game's instruction set reference and a task description, and generates a program in the game's language. The system prompt gives the complete ISA specification (opcodes, operand formats, addressing modes, memory layout, and execution semantics), ISA-level syntax snippets and programming patterns, and a response format that asks the model to reason inside \texttt{<think>} tags and return the program inside \texttt{<answer>} tags. It contains no worked task solutions. Results with 2--5 worked solutions in each prompt are reported in Appendix~\ref{app:results_examples}, and prompt templates are given in Appendix~\ref{app:prompts}.

Models are served with Ollama using each checkpoint's stored precision and quantization, a temperature of 0.7, and a maximum output length of 4096 tokens. The program is extracted from the answer tags, falling back to text after the reasoning delimiter, then to a Markdown code block, and finally to the raw response.

Models are given up to five attempts per task. On the first attempt, the model sees the task description and ISA reference, without worked task solutions in the system prompt. If the program fails, structured execution feedback is returned for the next attempt: a parse error if the code could not be read, a runtime error if execution halted, or per-test output comparisons if the program ran but produced incorrect results. A task is considered solved if a program passes all verification cases on any attempt within the budget; correctness is assessed on those cases. We report results under both single-shot evaluation (first attempt only) and iterative evaluation (all five attempts).

\section{Results}

Figure~\ref{fig:solve_rates} shows the number of tasks solved per game for each model, distinguishing first-attempt solves from those gained through iterative retries, and Table~\ref{tab:model_results} gives the per-model counts.

\begin{table}[t]
\centering
\setlength{\tabcolsep}{3pt}
\scriptsize
\begin{tabular}{llcccccr}
\toprule
\textbf{Model} & \textbf{T} & \textbf{HRM} & \textbf{SIC-1} & \textbf{TIS} & \textbf{MHRD} & \textbf{BOX} & \textbf{Total} \\
 & & \textbf{(36)} & \textbf{(31)} & \textbf{(7)} & \textbf{(8)} & \textbf{(10)} & \textbf{(92)} \\
\midrule

Gemma-4 31B        & T & 18/32 & 2/13 & 1/3 & 2/8 & 0/0 & 23/56 \\
Qwen3 32B          & T & 7/16 & 1/4 & 1/2 & 4/7 & 0/0 & 13/29 \\
Qwen3 14B          & T & 8/13 & 2/2 & 1/2 & 3/7 & 0/0 & 14/24 \\
Phi-4-Reasoning    & T & 0/20 & 0/1 & 0/1 & 1/2 & 0/0 & 1/24 \\
DeepSeek-R1 32B    & T & 8/13 & 0/2 & 1/1 & 3/7 & 0/0 & 12/23 \\
Devstral-Small-2   & C & 6/12 & 1/3 & 1/1 & 4/6 & 0/0 & 12/22 \\
Qwen3-Coder 30B    & C & 8/14 & 2/2 & 0/0 & 2/5 & 0/0 & 12/21 \\
Qwen3 8B           & T & 9/12 & 1/3 & 1/1 & 2/4 & 0/0 & 13/20 \\
Q2.5-Coder 32B     & C & 5/7 & 0/0 & 0/1 & 5/5 & 0/0 & 10/13 \\
Granite-4.1 8B     & G & 3/6 & 1/1 & 0/0 & 2/5 & 0/0 & 6/12 \\
Q2.5-Coder 14B     & C & 2/3 & 0/1 & 0/1 & 1/2 & 0/0 & 3/7 \\
Phi-4 14B          & G & 1/1 & 0/1 & 0/1 & 2/3 & 0/0 & 3/6 \\
Codestral 22B      & C & 1/2 & 0/1 & 0/0 & 2/3 & 0/0 & 3/6 \\
DS-Coder-V2 16B    & C & 0/3 & 0/0 & 0/0 & 2/2 & 0/0 & 2/5 \\

\bottomrule
\end{tabular}
\caption{
Per-model solve counts reported as solve@1 / solve@5,
corresponding respectively to first-attempt performance and
performance after up to five iterative attempts with execution
feedback. Task counts per environment are shown in parentheses.
Model type abbreviations: T = thinking/reasoning model,
C = coder-specialized model, G = general-purpose model.
}
\label{tab:model_results}
\end{table}

The best-performing model, Gemma-4 31B, has a reported total of 56 of 92 tasks (60.9\%). Nine of the 14 models solve fewer than 25\%. Performance varies substantially across games. Gemma-4 is the only model to complete all 8 MHRD tasks, while three others reach 7. HRM shows the widest spread in solved-task counts, ranging from 1 to 32. On SIC-1, Gemma-4 solves 13 of 31 tasks and Qwen3 32B solves 4; the other models solve 0--3. TIS-100 remains difficult, with no model solving more than 3 of 7 tasks. No model solves any BOX-256 task.

Retries yield large gains for some models, extending the execution-feedback setting studied in prior work~\cite{olausson2024selfrepair, chen2024selfdebug}. Phi-4-Reasoning improves from 0 to 20 solved HRM tasks and from 1 to 24 across all games. Gemma-4 gains 14 additional HRM solves and improves from 2 to 13 on SIC-1. On MHRD, Gemma-4 improves from 2 to 8, while Qwen3 14B and DeepSeek-R1 improve from 3 to 7. These changes show that additional attempts with feedback can accompany substantial improvements.

Thinking models as a group achieve a 31.9\% solve rate, compared to 13.4\% for code-specialized models and 9.8\% for general-purpose models (Table~\ref{tab:model_results}). Within the Qwen3 family, solve rate increases from 8B (21.7\%) to 14B (26.1\%) to 32B (31.5\%), while remaining below Gemma-4's reported performance.

We assess run-to-run variability over three runs for Gemma-4 31B, Devstral-Small-2, and Qwen2.5-Coder 32B on HRM, SIC-1, TIS-100, and MHRD. Each run allows up to five attempts per task. Table~\ref{tab:variance} summarizes first-attempt performance (solve@1) and performance within the full attempt budget (solve@5).

\begin{table}[t]
\centering
\scriptsize
\setlength{\tabcolsep}{1.5pt}
\renewcommand{\arraystretch}{1.15}
\begin{tabular}{@{}llcccc@{}}
\toprule
\textbf{Model} & \textbf{Metric} & \textbf{HRM} & \textbf{SIC-1} & \textbf{TIS} & \textbf{MHRD} \\
 & & \textbf{(36)} & \textbf{(31)} & \textbf{(7)} & \textbf{(8)} \\
\midrule
Gemma-4 31B      & solve@1 & $17.7 \pm 2.1$ & $2.0 \pm 1.0$ & $1.3 \pm 0.6$ & $2.3 \pm 0.6$ \\
                 & solve@5 & $32.0 \pm 1.0$ & $13.0 \pm 2.0$ & $2.7 \pm 0.6$ & $7.7 \pm 0.6$ \\
\midrule
Devstral-S-2 & solve@1 & $5.7 \pm 2.1$  & $1.0 \pm 1.0$ & $0.7 \pm 0.6$ & $3.7 \pm 1.5$ \\
                 & solve@5 & $11.7 \pm 2.1$ & $3.0 \pm 1.0$ & $1.0 \pm 0.0$ & $6.0 \pm 1.0$ \\
\midrule
Q2.5-Coder 32B & solve@1 & $5.3 \pm 1.2$ & $0.0 \pm 0.0$ & $0.3 \pm 0.6$ & $4.7 \pm 0.6$ \\
                  & solve@5 & $7.0 \pm 1.0$ & $0.0 \pm 0.0$ & $1.0 \pm 0.0$ & $5.3 \pm 0.6$ \\
\bottomrule
\end{tabular}
\caption{Solved-task counts across three runs, reported as mean $\pm$ sample standard deviation. Task counts per environment are shown in parentheses.}
\label{tab:variance}
\end{table}

\section{Failure Analysis}

Each attempt produces one of four outcomes, determined by the stage at which execution terminates:
\begin{itemize}[nosep,leftmargin=*]
\item \textbf{Parse error.} The program is not valid in the game's instruction set. The parser or assembler rejects it before execution begins.
\item \textbf{Runtime error.} The program parses but fails during execution, typically due to an infinite loop, illegal memory access, or stack overflow.
\item \textbf{Wrong output.} The program runs to completion but produces incorrect results on one or more test cases.
\item \textbf{Solved.} All test cases pass.
\end{itemize}

\noindent Figure~\ref{fig:failure_breakdown} shows the first-attempt outcome distributions by game, aggregated across all models.

\begin{table*}[t]
\centering
\small
\setlength{\tabcolsep}{5pt}
\renewcommand{\arraystretch}{1.25}
\begin{tabularx}{\textwidth}{@{}c X X@{}}
\toprule
\textbf{Score} & \textbf{Plan correctness} & \textbf{Implementation correctness} \\
\midrule
0 & Incorrect or irrelevant reasoning; the model does not identify a viable computational strategy & Non-executable output; parser failure, reasoning text leaked into code block, or wrong output format \\
1 & Partial reasoning that touches relevant concepts but misses the core algorithmic structure & Code that parses but is structurally broken; wrong opcodes, missing control flow, or fundamentally flawed logic \\
2 & Mostly correct high-level strategy with incomplete or flawed low-level details & Near-correct implementation with localized bugs such as off-by-one errors, wrong addresses, or missing edge cases \\
3 & Correct task decomposition and computational strategy for the target ISA & Correct or nearly correct execution; all or almost all test cases pass \\
\bottomrule
\end{tabularx}
\caption{Scoring rubric for the reasoning--execution gap (REG) analysis. For each task, each judge independently scores the model's computational plan and its implementation under this fixed rubric. Solved tasks receive 3/3 automatically; unsolved tasks are judged from the model's final attempt including its reasoning trace, generated code, and execution feedback.}
\label{tab:rubric}
\end{table*}

For SIC-1, parse errors account for 88.71\% of first attempts, runtime errors for 4.15\%, wrong outputs for 4.84\%, and solves for 2.30\%. The reported parse-error shares are 72.22\% for HRM, 47.96\% for TIS-100, 41.07\% for MHRD, and 81.43\% for BOX-256. TIS-100's reported wrong-output share is 43.88\%.

Table~\ref{tab:parse_causes} divides first-attempt parse errors by cause.

\begin{table}[t]
\centering
\footnotesize
\setlength{\tabcolsep}{5pt}
\setlength{\tabcolsep}{5pt}
\begin{tabular}{@{}lccc@{}}
\toprule
\textbf{Environment} & \textbf{No program} & \textbf{Formatting} & \textbf{ISA syntax} \\
 & \textbf{(\%)} & \textbf{(\%)} & \textbf{(\%)} \\
\midrule
SIC-1   & 37.1 & 17.7 & 45.2 \\
HRM     & 41.7 & 5.2  & 53.1 \\
TIS-100 & 32.5 & 2.5  & 65.0 \\
MHRD    & 3.2  & 9.7  & 87.1 \\
BOX-256 & 46.5 & 6.1  & 47.4 \\
\midrule
All     & 38.6 & 10.7 & 50.7 \\
\bottomrule
\end{tabular}
\caption{First-attempt parse errors by cause. \emph{No program} denotes responses in which the model exhausted its output budget while reasoning and returned no program, so the parser received reasoning text. \emph{Formatting} denotes programs that failed to follow the required answer format, such as code in Markdown fences. \emph{ISA syntax} denotes programs rejected by the parser for invalid opcodes, wrong operand counts or ranges, undefined or duplicate labels, unsupported comment syntax, or nonexistent components.}
\label{tab:parse_causes}
\end{table}

\section{Reasoning--Execution Gap Analysis}
\label{sec:reg_analysis}

The execution-level taxonomy above identifies where generated programs fail, but it does not distinguish conceptual reasoning failures from low-level implementation failures. A program classified as ``wrong output'' may arise because the model misunderstood the computational structure of the task, or because it inferred the correct strategy but failed during implementation. To separate these, we perform a structured reasoning--execution analysis.

For each model--task pair, we select the first attempt that passes all verification cases; if no attempt passes, we select the final attempt. In a separate diagnostic evaluation, GPT-5.5 and Claude Opus 4.8 each scored the final attempt for each unsolved task along two dimensions under a fixed rubric: \textbf{plan correctness} (did the model identify the right computational strategy?) and \textbf{implementation correctness} (did the code correctly execute that strategy?). Both use a 0--3 scale, summarized in Table~\ref{tab:rubric}. Each judge received the model's reasoning trace, generated code, and execution feedback, and scored each dimension independently (full judge prompt in Appendix~\ref{app:judge_prompt}; representative judgments in Appendix~\ref{app:reg_examples}).

Each judge scores both the plan and the implementation of each attempt, so the two scores share one calibration and their difference is not affected by differences between judges. Tasks solved by the verifier receive 3/3, since the verifier already establishes that the program implements a correct strategy. Unsolved tasks are judged from the final attempt. Environment-level comparisons involve only five environments and are therefore descriptive. We treat REG as an exploratory diagnostic that complements the solve rates.

We define the \emph{reasoning--execution gap} (REG) as
\[
\mathrm{REG} = \mathrm{PlanScore} - \mathrm{ImplementationScore}.
\]
High REG values indicate cases where models infer a plausible computational strategy but fail to express it as executable code in the target ISA.

Table~\ref{tab:reg_summary_mean} reports mean plan correctness, implementation correctness, and REG averaged across all 14 models for each environment under both judges (per-model GPT-5.5 scores in Appendix Table~\ref{tab:reg_summary_full}). Both judges assign BOX-256 the largest gap (0.96 and 1.92), indicating that models frequently identify a plausible spatial strategy but cannot translate it into valid programs. The judges differ on the remaining environments. GPT-5.5 places SIC-1 next (0.94), while Claude Opus 4.8 places HRM next (1.36). The two judges rank the five environments identically by implementation score but differently by plan score (Spearman 0.40), and Claude Opus 4.8 assigns higher plan scores overall (mean 2.52 vs.\ 1.96), which yields larger gaps.

MHRD has the highest mean plan and implementation scores and the smallest gap under both judges (0.06 and 0.38), indicating close alignment between these scores in the diagnostic evaluation. The compositional structure of NAND-based design is one possible explanation, but the aggregate scores do not establish why this environment differs from the others.

\begin{table}[t]
\centering
\footnotesize
\setlength{\tabcolsep}{3.5pt}
\begin{tabular}{@{}lcccccc@{}}
\toprule
 & \multicolumn{3}{c}{\textbf{GPT-5.5}} & \multicolumn{3}{c}{\textbf{Claude Opus 4.8}} \\
\cmidrule(lr){2-4}\cmidrule(l){5-7}
\textbf{Environment} & P & I & G & P & I & G \\
\midrule
BOX-256 & 1.44 & 0.48 & 0.96 & 2.83 & 0.91 & 1.92 \\
SIC-1   & 1.71 & 0.77 & 0.94 & 1.97 & 1.01 & 0.96 \\
HRM     & 2.07 & 1.26 & 0.81 & 2.79 & 1.43 & 1.36 \\
TIS-100 & 1.97 & 1.13 & 0.84 & 2.08 & 1.07 & 1.01 \\
MHRD    & 2.60 & 2.54 & 0.06 & 2.91 & 2.53 & 0.38 \\
\bottomrule
\end{tabular}
\caption{Mean plan correctness (P), implementation correctness (I), and reasoning--execution gap ($G=P-I$) averaged across all 14 models, scored by GPT-5.5 and Claude Opus 4.8. Higher G indicates models plan correctly but fail during implementation. Per-model GPT-5.5 scores are in Appendix Table~\ref{tab:reg_summary_full}.}
\label{tab:reg_summary_mean}
\end{table}

Associations between REG and other quantities depend on the judge. The rank correlation between first-attempt parse-error rates (Figure~\ref{fig:failure_breakdown}) and mean REG across the five environments is 0.80 under GPT-5.5 and 0.40 under Claude Opus 4.8. Excluding MHRD, code-specialized models have larger mean gaps than thinking models under GPT-5.5 (1.02 vs.\ 0.56) but smaller gaps under Claude Opus 4.8 (0.90 vs.\ 1.83).

\section{Qualitative Examples}

We show two solves and one failure below (more in Appendix~\ref{app:examples}).

\begingroup\scriptsize
\begin{tcolorbox}[colback=clrMHRD!10, colframe=clrMHRD!80!black, coltitle=white, title={\small\bfseries MHRD: NOT gate (solved)}, fontupper=\small, boxrule=0.8pt, top=1pt, bottom=1pt, left=2pt, right=2pt]
\textbf{Task:} Implement NOT from NAND gates.
\textbf{Model:} Gemma-4 31B (1st attempt)
{\scriptsize\begin{verbatim}
Inputs: in;  Outputs: out;
Parts: n1 NAND;
Wires: in -> n1.in1, in -> n1.in2,
       n1.out -> out;
\end{verbatim}
}%
\vspace{-4pt}
\small $\mathrm{NAND}(x,x) = \overline{x \land x} = \overline{x}$. The standard single-gate NOT.
\end{tcolorbox}
\vspace{2pt}
\begin{tcolorbox}[colback=clrTIS!10, colframe=clrTIS, coltitle=white, title={\small\bfseries TIS-100: Signal Amplifier (solved)}, fontupper=\small, boxrule=0.8pt, top=1pt, bottom=1pt, left=2pt, right=2pt]
\textbf{Task:} Double each input value.
\textbf{Model:} Gemma-4 31B (1st attempt)
{\scriptsize\begin{verbatim}
@0                @4              @8
MOV UP, ACC       MOV UP, DOWN    MOV UP, DOWN
ADD ACC
MOV ACC, DOWN
\end{verbatim}
}%
\vspace{-4pt}
\small Node~0 doubles; nodes~4,~8 relay to output. All run in parallel, synchronized by blocking reads.
\end{tcolorbox}
\vspace{2pt}
\begin{tcolorbox}[colback=clrBOX!15, colframe=black!60, coltitle=white, title={\small\bfseries BOX-256: Rectangle Frame (failed)}, fontupper=\small, boxrule=0.8pt, top=1pt, bottom=1pt, left=2pt, right=2pt]
\textbf{Task:} Draw a rectangle frame on the 16$\times$16 grid.
\textbf{Model:} Gemma-4 31B (5 attempts, 0/1 tests passed)\\
\textbf{Error:} \texttt{Unknown instruction: <THINK>}\\[2pt]
The model exhausted its output budget while reasoning and never opened an answer block, so the parser received the reasoning text.
\end{tcolorbox}
\endgroup

\section{Iterative Refinement and Retry Behavior}

Figure~\ref{fig:solve_rates} reports solve@1 alongside solve@5 (per-model counts in Table~\ref{tab:model_results}). Across all model--task pairs, the reported totals increase from 76 to 154 on HRM, 10 to 34 on SIC-1, 6 to 14 on TIS-100, and 35 to 66 on MHRD. BOX-256 remains unsolved. Absolute gains are largest on HRM and MHRD, although SIC-1 also shows a substantial gain for Gemma-4. The aggregate counts do not identify which gains result from syntax repair, changes in computational strategy, or additional sampling.

\section{Conclusion}

ISA-Bench provides programming-game environments with full execution stacks. Thinking models have higher average solve rates in this evaluation, and additional attempts with execution feedback yield uneven gains. The REG analysis provides a complementary view of planning and implementation scores. Together, these results highlight the gap between forming a plausible computational strategy and expressing it as a correct program under the target ISA. ISA-Bench also provides a testbed for methods aimed at unfamiliar instruction sets, such as rule learning, program induction, and retrieval of ISA documentation.

\section*{Limitations}

All five environments use program synthesis with deterministic execution and fixed test cases, which does not capture interactive software systems or multi-file programming scenarios. Our model selection covers open-weight models between 8B and 32B parameters; larger or proprietary models may exhibit different failure profiles, though our focus on open-weight models is a deliberate design choice to enable reproducibility and downstream research (Section~4.1). Comparisons between model groups do not control for model size, training data, or inference behavior. The current game selection covers minimal-instruction arithmetic, sequential register planning, parallel message-passing, combinational circuit design, and spatial memory-mapped output. Notable omissions include stack-based architectures, interrupt-driven execution, and vector-parallel (SIMD) computation. The REG analysis uses two LLM judges under a fixed rubric, and some of its conclusions differ between them. Some game-adjacent content may appear in pretraining corpora.

\section*{LLM Usage Statement}

We used Claude Code and GitHub Copilot throughout this work. Under the authors' direction, they assisted with implementing the benchmark environments and evaluation scripts, editing the paper, and \LaTeX{} and figure scripts. GPT-5.5 and Claude Opus 4.8 were used as judges in the REG analysis. The authors reviewed all AI-generated code and text, verified all citations, and take full responsibility for the content. All numbers were taken from evaluation script outputs and not fabricated or hallucinated by AI agents.

\bibliography{references}

\clearpage

\appendix
\raggedbottom

\renewcommand{\contentsname}{Appendix Contents}
\begingroup
\let\clearpage\relax
\tableofcontents
\endgroup
\smallskip

\section{Per-Game ISA Reference}
\label{app:isa}

Each game's instruction set is summarized below. Detailed opcode descriptions, addressing modes, and execution semantics are provided in the full system prompts (Appendix~\ref{app:prompts_full}).

\paragraph{SIC-1.} One instruction: \texttt{subleq A, B [, C]}. Subtracts \texttt{mem[B]} from \texttt{mem[A]} and branches to \texttt{C} if the result is non-positive. Memory is 256 signed bytes. Special addresses: 253 (\texttt{@IN}), 254 (\texttt{@OUT}), 255 (\texttt{@HALT}). The assembler supports labels, \texttt{.data} directives, label offsets, and negated addresses.

\paragraph{HRM.} 11 opcodes: \texttt{INBOX}, \texttt{OUTBOX}, \texttt{COPYFROM}, \texttt{COPYTO}, \texttt{ADD}, \texttt{SUB}, \texttt{BUMPUP}, \texttt{BUMPDN}, \texttt{JUMP}, \texttt{JUMPZ}, \texttt{JUMPN}. Direct and indirect (\texttt{[n]}) addressing. Single accumulator register, indexed memory tiles. Values are integers ($-999$ to $999$) or single uppercase letters.

\paragraph{TIS-100.} 13 opcodes: \texttt{MOV}, \texttt{ADD}, \texttt{SUB}, \texttt{NEG}, \texttt{SAV}, \texttt{SWP}, \texttt{JMP}, \texttt{JEZ}, \texttt{JNZ}, \texttt{JGZ}, \texttt{JLZ}, \texttt{JRO}, \texttt{NOP}. Each node has \texttt{ACC} and \texttt{BAK} registers. Port communication (\texttt{UP}, \texttt{DOWN}, \texttt{LEFT}, \texttt{RIGHT}) is blocking. Programs are specified per node with \texttt{@N} headers. 15-instruction limit per node.

\paragraph{MHRD.} Four HDL keywords: \texttt{Inputs}, \texttt{Outputs}, \texttt{Parts}, \texttt{Wires}. The only primitive is a \texttt{NAND} gate with ports \texttt{in1}, \texttt{in2}, \texttt{out}. Circuits are composed by instantiating parts and wiring them together. Supports multi-bit buses (\texttt{in[4]}).

\paragraph{BOX-256.} 13 opcodes: \texttt{MOV}, \texttt{ADD}, \texttt{SUB}, \texttt{MUL}, \texttt{DIV}, \texttt{MOD}, \texttt{JMP}, \texttt{JEQ}, \texttt{JNE}, \texttt{JGR}, \texttt{PIX}, \texttt{FLP}, \texttt{THR}. Three addressing modes: immediate (\texttt{\#N}), direct (\texttt{@N}), indirect (\texttt{*N}). Memory is 256 bytes shared between code and display. Each instruction is 4 bytes. Programs run for up to 10{,}000 cycles.

\section{Full Task Lists}
\label{app:tasks}

\paragraph{SIC-1 (31 tasks).} Each task specifies an input/output transformation over integer sequences using only the SUBLEQ instruction. Tasks progress from basic I/O to self-hosting interpreters.
\begin{enumerate}[nosep,leftmargin=*,font=\small]
\item \textbf{Subleq Instruction and Output.} Negate each input value.
\item \textbf{Data Directive and Looping.} Negate inputs in a loop using data directives.
\item \textbf{First Assessment.} Echo each input unchanged.
\item \textbf{Addition.} Add pairs of inputs.
\item \textbf{Subtraction.} Subtract pairs of inputs.
\item \textbf{Sign Function.} Output $-1$, $0$, or $1$ based on the sign of each input.
\item \textbf{Multiplication.} Multiply pairs of inputs.
\item \textbf{Division.} Divide pairs of inputs, output quotient and remainder.
\item \textbf{Sequence Sum.} Sum a zero-terminated sequence.
\item \textbf{Sequence Cardinality.} Count elements in a zero-terminated sequence.
\item \textbf{Number to Sequence.} Expand a number $n$ into the sequence $1, 2, \ldots, n$.
\item \textbf{Self-Modifying Code.} Read instructions from input and execute them.
\item \textbf{Stack Memory.} Reverse a fixed-length sequence (LIFO).
\item \textbf{Reverse Sequence.} Reverse a zero-terminated sequence.
\item \textbf{Interleave.} Interleave two zero-terminated sequences.
\item \textbf{Indicator Function.} Output 1 if an element appears in a reference set, 0 otherwise.
\item \textbf{Sort.} Sort a zero-terminated sequence in ascending order.
\item \textbf{Mode.} Find the most frequent element in a zero-terminated sequence.
\item \textbf{Characters.} Output a hardcoded string as ASCII values.
\item \textbf{Decimal Digits.} Convert an ASCII digit character to its numeric value.
\item \textbf{Uppercase.} Convert lowercase ASCII letters to uppercase.
\item \textbf{Strings.} Output a hardcoded multi-character string.
\item \textbf{Tokenizer.} Split an ASCII string on spaces (replace spaces with zeros).
\item \textbf{Parse Decimal.} Parse a zero-terminated ASCII string into an integer.
\item \textbf{Print Decimal.} Convert an integer to a zero-terminated ASCII string.
\item \textbf{Calculator.} Parse and evaluate simple arithmetic expressions from ASCII.
\item \textbf{Multi-Line Strings.} Handle newline-separated strings.
\item \textbf{Parse Data Directives.} Parse \texttt{.data} assembler directives from ASCII.
\item \textbf{Parse Subleq Instructions.} Parse \texttt{subleq} instructions from ASCII.
\item \textbf{Self-Hosting.} Parse and execute a complete subleq program from input.
\item \textbf{Self-Hosting Part 2.} Self-hosting with data directives and labels.
\end{enumerate}

\paragraph{HRM (36 tasks).} Each task transforms an input conveyor (INBOX) to an output conveyor (OUTBOX).
\begin{enumerate}[nosep,leftmargin=*,font=\small]
\item \textbf{Mail Room.} Grab each thing from the INBOX and drop it into the OUTBOX.
\item \textbf{Busy Mail Room.} Grab each thing from the INBOX, and drop each one into the OUTBOX.
\item \textbf{Copy Floor.} Send specific letters from the floor tiles to the OUTBOX.
\item \textbf{Scrambler Handler.} Grab the first TWO things from the INBOX and drop them in reverse order.
\item \textbf{Rainy Summer.} For each two things in the INBOX, add them together.
\item \textbf{Zero Exterminator.} Send all things that ARE NOT ZERO to the OUTBOX.
\item \textbf{Tripler Room.} For each thing in the INBOX, TRIPLE it.
\item \textbf{Zero Preservation Initiative.} Send only the ZEROs to the OUTBOX.
\item \textbf{Octoplier Suite.} For each thing in the INBOX, multiply it by 8.
\item \textbf{Sub Hallway.} For each two things, subtract the 1st from the 2nd.
\item \textbf{Tetracontiplier.} For each thing in the INBOX, multiply it by 40.
\item \textbf{Equalization Room.} Get two things, send to OUTBOX if they are equal.
\item \textbf{Maximization Room.} Grab TWO things, put only the BIGGER in the OUTBOX.
\item \textbf{Absolute Positivity.} Send each thing, but remove negative signs.
\item \textbf{Exclusive Lounge.} Send 0 if same sign, 1 if different sign.
\item \textbf{Countdown.} For each number, send it followed by all numbers down to zero.
\item \textbf{Multiplication Workshop.} For each two things, multiply them.
\item \textbf{Zero Terminated Sum.} Sum each zero-terminated string.
\item \textbf{Fibonacci Visitor.} Send the Fibonacci sequence up to but not exceeding the input.
\item \textbf{The Littlest Number.} For each zero-terminated string, send only the smallest.
\item \textbf{Mod Module.} For each two things, output the remainder of dividing the first by the second.
\item \textbf{Cumulative Countdown.} Sum of the input plus all numbers down to zero.
\item \textbf{Small Divide.} How many times does the second fit into the first?
\item \textbf{Three Sort.} For each THREE things, send them sorted smallest to largest.
\item \textbf{Storage Floor.} Each input is an address; send the value at that floor tile.
\item \textbf{String Storage Floor.} Each input is an address of a zero-terminated string on the floor.
\item \textbf{String Reverse.} Reverse each zero-terminated string.
\item \textbf{Inventory Report.} Count matching items on the floor for each input.
\item \textbf{Vowel Incinerator.} Send everything except vowels.
\item \textbf{Duplicate Removal.} Send everything unless you have seen it before.
\item \textbf{Alphabetizer.} Determine which of two words comes first alphabetically.
\item \textbf{Scavenger Chain.} Follow a chain of floor tile addresses.
\item \textbf{Digit Exploder.} Send the individual digits of each number.
\item \textbf{Re-Coordinator.} Follow indirect address chains on the floor.
\item \textbf{Prime Factory.} Send prime factors in order from smallest to largest.
\item \textbf{Sorting Floor.} Sort each zero-terminated string, smallest first.
\end{enumerate}

\paragraph{TIS-100 (7 tasks).} Each task specifies input and output streams across a $4\times3$ node grid.
\begin{enumerate}[nosep,leftmargin=*,font=\small]
\item \textbf{Signal Amplifier.} Read values from IN.A, double them, write to OUT.A.
\item \textbf{Differential Converter.} Read A and B, output A-B and B-A.
\item \textbf{Signal Comparator.} Read input; output 1 to the column corresponding to its sign.
\item \textbf{Sequence Counter.} Read N, output 1, 2, 3, \ldots, N.
\item \textbf{Signal Edge Detector.} Output 1 when value increases, -1 when it decreases, 0 otherwise.
\item \textbf{Sequence Reverser.} Read two values, output them in reverse order.
\item \textbf{Signal Multiplier.} Read A and B, output $A \times B$ via repeated addition.
\end{enumerate}

\paragraph{MHRD (8 tasks).} Each task specifies a truth table to implement from NAND gates.
\begin{enumerate}[nosep,leftmargin=*,font=\small]
\item \textbf{NOT.} Invert the input signal.
\item \textbf{AND.} Output 1 only when both inputs are 1.
\item \textbf{OR.} Output 1 when at least one input is 1.
\item \textbf{XOR.} Output 1 when inputs are different.
\item \textbf{MUX.} Select in1 when sel=0, in2 when sel=1.
\item \textbf{DEMUX.} Route input to out1 when sel=0, out2 when sel=1.
\item \textbf{HALFADDER.} Add two 1-bit numbers, produce sum and carry.
\item \textbf{FULLADDER.} Add two 1-bit numbers with carry in, produce sum and carry out.
\end{enumerate}

\paragraph{BOX-256 (10 tasks).} Each task specifies a target pixel pattern on a $16\times16$ grid. Figure~\ref{fig:box256_targets} shows all 10 target patterns.
\begin{enumerate}[nosep,leftmargin=*,font=\small]
\item \textbf{Rectangle Frame.} Draw a thin rectangle frame.
\item \textbf{Thick Frame.} Draw a thick rectangle frame.
\item \textbf{Checkerboard.} Draw a 1$\times$1 checkerboard pattern.
\item \textbf{Big Checkerboard.} Draw a 2$\times$2 checkerboard pattern.
\item \textbf{Corner Squares.} Draw four hollow squares in corners.
\item \textbf{Triangle.} Draw a stepped triangle.
\item \textbf{Diagonals.} Draw diagonal stripes.
\item \textbf{Spiral.} Draw a spiral pattern.
\item \textbf{Sierpinski Triangle.} Draw a Sierpinski triangle.
\item \textbf{Colored Sierpinski.} Draw a colored Sierpinski triangle.
\end{enumerate}

\input{figures/fig_box256targets}

\section{Per-Model Results}
\label{app:results}

\begin{table*}[ht]
\centering
\footnotesize
\setlength{\tabcolsep}{3.2pt}
\begin{tabular}{lccccccccccccccc}
\toprule
& \multicolumn{3}{c}{BOX-256}
& \multicolumn{3}{c}{HRM}
& \multicolumn{3}{c}{MHRD}
& \multicolumn{3}{c}{SIC-1}
& \multicolumn{3}{c}{TIS-100} \\
\cmidrule(lr){2-4}
\cmidrule(lr){5-7}
\cmidrule(lr){8-10}
\cmidrule(lr){11-13}
\cmidrule(lr){14-16}
Model
& P & I & G
& P & I & G
& P & I & G
& P & I & G
& P & I & G \\
\midrule

Gemma-4 31B
& 1.90 & 0.90 & 1.00
& 2.72 & 2.72 & 0.00
& 3.00 & 3.00 & 0.00
& 1.68 & 1.29 & 0.39
& 2.57 & 1.86 & 0.71 \\

Qwen3-32B
& 1.10 & 0.30 & 0.80
& 1.89 & 1.39 & 0.50
& 3.00 & 3.00 & 0.00
& 1.32 & 0.68 & 0.65
& 1.57 & 1.00 & 0.57 \\

Qwen3-14B
& 1.00 & 0.30 & 0.70
& 1.83 & 1.36 & 0.47
& 3.00 & 3.00 & 0.00
& 1.26 & 0.71 & 0.55
& 1.57 & 1.29 & 0.29 \\

Qwen3-Coder 30B
& 1.90 & 0.70 & 1.20
& 2.31 & 1.50 & 0.81
& 3.00 & 3.00 & 0.00
& 2.16 & 1.42 & 0.74
& 2.29 & 1.57 & 0.71 \\

DeepSeek-R1 32B
& 1.00 & 0.60 & 0.40
& 1.83 & 1.42 & 0.42
& 2.75 & 2.75 & 0.00
& 1.16 & 0.65 & 0.52
& 1.57 & 1.14 & 0.43 \\

Devstral-Small-2
& 1.40 & 1.00 & 0.40
& 2.25 & 1.75 & 0.50
& 2.38 & 2.25 & 0.13
& 2.00 & 1.13 & 0.87
& 2.29 & 1.57 & 0.71 \\

Qwen3-8B
& 1.00 & 0.20 & 0.80
& 1.67 & 1.14 & 0.53
& 2.75 & 2.75 & 0.00
& 1.13 & 0.35 & 0.77
& 1.57 & 0.86 & 0.71 \\

Phi-4-Reasoning
& 1.00 & 0.10 & 0.90
& 1.72 & 1.72 & 0.00
& 2.25 & 2.25 & 0.00
& 1.13 & 0.23 & 0.90
& 1.57 & 1.14 & 0.43 \\

Q2.5-Coder 32B
& 1.70 & 0.30 & 1.40
& 2.19 & 0.89 & 1.31
& 2.38 & 2.25 & 0.13
& 2.06 & 0.97 & 1.10
& 2.14 & 1.00 & 1.14 \\

Q2.5-Coder 14B
& 1.70 & 0.80 & 0.90
& 2.22 & 0.89 & 1.33
& 2.50 & 2.50 & 0.00
& 2.13 & 1.03 & 1.10
& 2.29 & 1.43 & 0.86 \\

Granite-4.1 8B
& 2.00 & 0.20 & 1.80
& 2.22 & 1.06 & 1.17
& 2.63 & 2.25 & 0.38
& 2.06 & 0.55 & 1.52
& 2.00 & 0.86 & 1.14 \\

Phi-4 14B
& 1.90 & 0.40 & 1.50
& 2.19 & 0.58 & 1.61
& 2.50 & 2.38 & 0.13
& 2.03 & 0.45 & 1.58
& 2.14 & 0.71 & 1.43 \\

Codestral 22B
& 1.30 & 0.60 & 0.70
& 1.92 & 0.69 & 1.22
& 2.25 & 2.25 & 0.00
& 1.90 & 0.68 & 1.23
& 2.00 & 0.57 & 1.43 \\

DS-Coder-V2 16B
& 1.30 & 0.30 & 1.00
& 2.03 & 0.58 & 1.44
& 2.00 & 1.88 & 0.13
& 1.90 & 0.68 & 1.23
& 2.00 & 0.86 & 1.14 \\

\midrule

\textbf{Mean}
& \textbf{1.44} & \textbf{0.48} & \textbf{0.96}
& \textbf{2.07} & \textbf{1.26} & \textbf{0.81}
& \textbf{2.60} & \textbf{2.54} & \textbf{0.06}
& \textbf{1.71} & \textbf{0.77} & \textbf{0.94}
& \textbf{1.97} & \textbf{1.13} & \textbf{0.84} \\

\bottomrule
\end{tabular}
\caption{
Per-model plan correctness (P), implementation correctness (I), and reasoning--execution gap (G = P $-$ I) across all 14 models and five environments. Mean row corresponds to the GPT-5.5 columns of Table~\ref{tab:reg_summary_mean} in the main text.
}
\label{tab:reg_summary_full}
\end{table*}

\FloatBarrier
\subsection{Results with worked examples}
\label{app:results_examples}

Table~\ref{tab:model_results_examples} reports results with 2--5 worked task solutions included in each system prompt. All other settings follow Section~\ref{sec:eval_protocol}.

\begin{table}[ht]
\centering
\setlength{\tabcolsep}{3pt}
\scriptsize
\begin{tabular}{llcccccr}
\toprule
\textbf{Model} & \textbf{T} & \textbf{HRM} & \textbf{SIC-1} & \textbf{TIS} & \textbf{MHRD} & \textbf{BOX} & \textbf{Total} \\
 & & \textbf{(36)} & \textbf{(31)} & \textbf{(7)} & \textbf{(8)} & \textbf{(10)} & \textbf{(92)} \\
\midrule
Gemma-4 31B        & T & 16/28 & 0/1 & 2/4 & 6/8 & 0/0 & 24/41 \\
Qwen3 32B          & T & 9/16 & 2/5 & 2/2 & 2/8 & 0/0 & 15/31 \\
Qwen3 14B          & T & 13/15 & 3/4 & 1/2 & 3/8 & 0/0 & 20/29 \\
Qwen3-Coder 30B    & C & 8/11 & 3/7 & 1/2 & 3/8 & 0/0 & 15/28 \\
DeepSeek-R1 32B    & T & 12/15 & 2/2 & 2/2 & 7/7 & 0/0 & 23/26 \\
Devstral-Small-2   & C & 12/15 & 2/4 & 2/2 & 5/5 & 0/0 & 21/26 \\
Qwen3 8B           & T & 11/12 & 1/2 & 2/2 & 5/7 & 0/0 & 19/23 \\
Phi-4-Reasoning    & T & 2/13 & 1/2 & 2/2 & 4/5 & 0/0 & 9/22 \\
Q2.5-Coder 32B     & C & 7/10 & 2/4 & 2/2 & 5/6 & 0/0 & 16/22 \\
Q2.5-Coder 14B     & C & 8/8 & 2/3 & 1/1 & 3/5 & 0/0 & 14/17 \\
Granite-4.1 8B     & G & 7/9 & 2/2 & 0/0 & 5/5 & 0/0 & 14/16 \\
Phi-4 14B          & G & 6/7 & 1/1 & 1/1 & 6/6 & 0/0 & 14/15 \\
Codestral 22B      & C & 7/7 & 1/1 & 1/1 & 2/5 & 0/0 & 11/14 \\
DS-Coder-V2 16B    & C & 5/5 & 1/1 & 1/1 & 4/4 & 0/0 & 11/11 \\
\bottomrule
\end{tabular}
\caption{Per-model solve@1 / solve@5 counts with worked task solutions included in the system prompt. Abbreviations follow Table~\ref{tab:model_results}.}
\label{tab:model_results_examples}
\end{table}

\FloatBarrier
\section{Worked Examples}
\label{app:examples}

Illustrative outputs are shown below, separately from the system prompts used for evaluation. BOX-256 shows a representative failure.

\begin{tcolorbox}[colback=clrHRM!15, colframe=clrHRM!80!black, coltitle=white, title={\small\bfseries HRM: Mail Room}, fontupper=\small, boxrule=0.8pt]
\textbf{Task:} Grab each thing from the INBOX and drop it into the OUTBOX.\\
\textbf{Model:} Gemma-4 31B (first attempt, 2/2 tests passed)
\begin{verbatim}
loop:
    INBOX     ; grab from input
    OUTBOX    ; send to output
    JUMP loop ; repeat
\end{verbatim}
\smallskip
\small The simplest possible HRM program. \texttt{INBOX} implicitly halts when the input stream is exhausted, so no explicit termination condition is needed.
\end{tcolorbox}

\begin{tcolorbox}[colback=clrSIC!10, colframe=clrSIC, coltitle=white, title={\small\bfseries SIC-1: Subleq Instruction and Output}, fontupper=\small, boxrule=0.8pt]
\textbf{Task:} Negate the supplied input.\\
\textbf{Model:} Qwen3 14B (first attempt, 1/1 tests passed)
\begin{verbatim}
subleq @OUT, @IN, @HALT
\end{verbatim}
\smallskip
\small For the test input 3, the instruction computes $0 - 3$, outputs $-3$, and branches to the halt address.
\end{tcolorbox}

\begin{tcolorbox}[colback=clrTIS!10, colframe=clrTIS, coltitle=white, title={\small\bfseries TIS-100: Signal Amplifier}, fontupper=\small, boxrule=0.8pt]
\textbf{Task:} Read values from IN.0, double them, write to OUT.0.\\
\textbf{Model:} Gemma-4 31B (first attempt, 1/1 tests passed)
\begin{verbatim}
@0             ; node 0 (input)
MOV UP, ACC    ; read input
ADD ACC        ; double it
MOV ACC, DOWN  ; send down

@4             ; node 4 (relay)
MOV UP, DOWN

@8             ; node 8 (output)
MOV UP, DOWN
\end{verbatim}
\smallskip
\small Data flows vertically: node~0 reads and doubles, nodes~4 and~8 relay to output. Each program loops implicitly. All three execute in parallel, synchronized by blocking port reads.
\end{tcolorbox}

\begin{tcolorbox}[colback=clrMHRD!10, colframe=clrMHRD!80!black, coltitle=white, title={\small\bfseries MHRD: NOT}, fontupper=\small, boxrule=0.8pt]
\textbf{Task:} Implement a NOT gate from NAND gates.\\
\textbf{Model:} Gemma-4 31B (first attempt, 2/2 tests passed)
\begin{verbatim}
Inputs: in;
Outputs: out;
Parts: n1 NAND;  ; one NAND gate
Wires:
  in -> n1.in1,  ; tie both inputs
  in -> n1.in2,  ; NAND(x,x)=NOT(x)
  n1.out -> out;
\end{verbatim}
\smallskip
\small $\text{NAND}(x, x) = \overline{x \land x} = \overline{x}$. The standard single-gate NOT construction.
\end{tcolorbox}

\begin{tcolorbox}[colback=clrBOX!15, colframe=black!60, coltitle=white, title={\small\bfseries BOX-256: Rectangle Frame (failed)}, fontupper=\small, boxrule=0.8pt]
\textbf{Task:} Draw a rectangle frame on the 16$\times$16 grid.\\
\textbf{Model:} Gemma-4 31B (5 attempts, 0/1 tests passed)\\
\textbf{Error:} \texttt{Unknown instruction: <THINK> (line 1)}\\[4pt]
The model emitted its reasoning trace as literal text in the code block rather than executable BOX-256 assembly. The parser rejected the very first line.
\end{tcolorbox}


\section{Prompt Templates}
\label{app:prompts}

Each model receives a system prompt containing the game's full ISA specification and a response format instruction, without worked task solutions. Syntax snippets and ISA-level programming patterns remain part of the reference. The prompt structure is identical across all five games:

\begin{enumerate}[nosep]
\item \textbf{Role and task framing.} A single sentence establishing the model as an expert programmer for the game.
\item \textbf{ISA specification.} The complete instruction set reference: opcodes, operand formats, addressing modes, memory layout, and execution semantics.
\item \textbf{Response format.} Instructions to reason in \texttt{<think>} tags and provide the solution in \texttt{<answer>} tags.
\end{enumerate}

On failure, the model receives structured feedback appended to the conversation. Parse errors include the error message and line number. Runtime errors include the error type. Wrong-output failures include per-test-case comparisons of expected and actual output.

\section{Reward Functions}
\label{app:rewards}

Each environment provides a graded reward between 0 and 1, decomposed into a correctness component and an efficiency component:
\begin{align*}
r = {} & w_c \cdot \frac{\text{tests passed}}{\text{tests total}} \\
& + w_e \cdot \mathbb{1}[\text{all tests pass}] \cdot \text{efficiency}
\end{align*}
where $w_c = 0.7$ and $w_e = 0.3$ for all games except BOX-256, which uses $w_c = 0.8$ and $w_e = 0.2$. The efficiency bonus is awarded only when all test cases pass. Game-specific efficiency metrics are:

\begin{itemize}[nosep,leftmargin=*]
\item \textbf{SIC-1.} Full efficiency bonus for any correct solution (no par values).
\item \textbf{HRM.} Average of instruction efficiency ($\text{par\_size} / \text{actual\_size}$) and step efficiency ($\text{par\_speed} / \text{actual\_steps}$), each capped at 1.0.
\item \textbf{TIS-100.} Node efficiency: $1 - (\text{nodes\_used} - 1) / 11$, rewarding solutions that use fewer of the 12 available nodes.
\item \textbf{MHRD.} NAND efficiency: $\text{reference\_count} / \text{actual\_count}$, capped at 1.0, rewarding circuits that use fewer NAND gates.
\item \textbf{BOX-256.} Pixel accuracy as the base metric (partial credit for near-correct patterns). Size efficiency bonus: $1 - \text{code\_size} / 256$ for perfect solutions.
\end{itemize}

\section{Compute Resources}
\label{app:compute}

The original evaluation used a server with 8 NVIDIA A40 GPUs (48 GB each), with one Ollama instance per GPU. The reported approximately 72-hour inference time covers that original evaluation of 14 models across 92 tasks, with up to five attempts per task. No model training or fine-tuning was performed; all models were evaluated using publicly available weights. The reasoning--execution gap judgments were obtained via the GPT-5.5 and Claude Opus 4.8 APIs.

Inference settings are given in Section~\ref{sec:eval_protocol}.

\section{System Prompts}
\label{app:prompts_full}

The complete text of all five system prompts follows.

\input{appendix_prompts}

\section{LLM-as-Judge Prompt}
\label{app:judge_prompt}

The following prompt was provided to both judges to judge the final attempt for each unsolved task. The judge received the model's reasoning trace, generated code, and execution feedback alongside this prompt.

\begin{tcolorbox}[breakable, colback=black!4, colframe=black!40, boxrule=0.5pt, fontupper=\small]
\textbf{Task:} Plan--Execution Judgment

You are evaluating ISA-Bench model outputs.

\medskip
\textbf{Scores}

\texttt{plan\_score}:
\begin{itemize}[nosep,leftmargin=*]
\item 0 = wrong or irrelevant reasoning
\item 1 = partial useful reasoning but misses core algorithm
\item 2 = mostly correct high-level strategy but confused/incomplete details
\item 3 = correct task decomposition and computational strategy
\end{itemize}

\texttt{implementation\_score}:
\begin{itemize}[nosep,leftmargin=*]
\item 0 = no executable code, parser failure, reasoning leaked into code
\item 1 = executable-looking but structurally broken
\item 2 = close implementation with localized bug
\item 3 = correct or nearly correct implementation
\end{itemize}

\texttt{reasoning\_execution\_gap} = \texttt{plan\_score} $-$ \texttt{implementation\_score}

\medskip
\textbf{Failure categories:} \texttt{solved}, \texttt{wrong\_plan}, \texttt{correct\_plan\_bad\_execution}, \texttt{partial\_plan\_partial\_execution}, \texttt{syntax\_or\_format\_failure}, \texttt{control\_flow\_failure}, \texttt{address\_or\_constant\_bug}, \texttt{lucky\_or\_unjustified\_code}

\medskip
\textbf{Rules}
\begin{itemize}[nosep,leftmargin=*]
\item Judge the reasoning separately from the final code.
\item Do not reward verbosity.
\item If the reasoning correctly describes the solution but the code fails due to syntax, jump targets, memory addresses, or formatting, use \texttt{correct\_plan\_bad\_execution}.
\item If \texttt{<think>} or natural language appears inside executable code, \texttt{implementation\_score} must be 0.
\item If solved is true, \texttt{implementation\_score} should usually be 3.
\item If the code is correct but reasoning is absent or unjustified, use \texttt{lucky\_or\_unjustified\_code}.
\end{itemize}
\end{tcolorbox}

\section{LLM-as-Judge Examples}
\label{app:reg_examples}

The following examples illustrate representative reasoning--execution gap (REG) judgments from the separate diagnostic evaluation. Each example shows the model's reasoning, submitted code, execution outcome, and the judge's verdict explaining how the plan and implementation scores were assigned.

\begin{tcolorbox}[breakable, colback=black!8, colframe=black!40, coltitle=black, title={\small\bfseries Example 1: TIS-100 --- Differential Converter \hfill Plan: 2 \quad Impl: 0}, fontupper=\small, boxrule=0.8pt]
\textbf{Model:} Codestral 22B \quad \textbf{Category:} Correct plan, failed execution

\begin{tcolorbox}[colback=white, colframe=black!20, coltitle=black, boxrule=0.5pt, fontupper=\small, title={\scriptsize\bfseries Model reasoning}]
Node 0 reads A and sends it to node 1. At the same time, it keeps a copy for subtraction. Node 1 reads B, receives A from left, computes both differences, and sends them to output nodes.
\end{tcolorbox}

\begin{tcolorbox}[colback=white, colframe=black!20, coltitle=black, boxrule=0.5pt, fontupper=\small, title={\scriptsize\bfseries Submitted code}]
\begin{verbatim}
@0
MOV UP, ACC
SAV
MOV ACC, RIGHT
JMP LOOP_START
LOOP_START:
  MOV UP, ACC
  SWP
  SUB ACC
  MOV ACC, DOWN
  MOV RIGHT, LEFT
  JMP LOOP_START
@1
MOV UP, ACC
SAV
MOV ACC, LEFT
MOV LEFT, ACC
SWP
SUB ACC
MOV ACC, DOWN
JMP LOOP_START
...
\end{verbatim}
\end{tcolorbox}

\textbf{Execution:} \texttt{Node 1: Line 7: Undefined label: LOOP\_START}

\begin{tcolorbox}[colback=white, colframe=black!20, coltitle=black, boxrule=0.5pt, fontupper=\small, title={\scriptsize\bfseries Judge verdict}]
\textbf{Plan 2:} The reasoning correctly identifies the core communication pattern---route A and B to where both are available, compute both A$-$B and B$-$A---which is the right high-level strategy for a differential converter on a parallel grid.

\textbf{Impl 0:} The program never executes. A jump target (\texttt{LOOP\_START}) referenced in node~1 is never defined, causing a parser failure before any instruction runs.
\end{tcolorbox}
\end{tcolorbox}

\begin{tcolorbox}[breakable, colback=black!8, colframe=black!40, coltitle=black, title={\small\bfseries Example 2: HRM --- Absolute Positivity \hfill Plan: 1 \quad Impl: 1}, fontupper=\small, boxrule=0.8pt]
\textbf{Model:} DeepSeek-Coder-V2 16B \quad \textbf{Category:} Wrong plan

\begin{tcolorbox}[colback=white, colframe=black!20, coltitle=black, boxrule=0.5pt, fontupper=\small, title={\scriptsize\bfseries Model reasoning}]
If a number is negative, we should remove its negative sign by converting it to positive. We can do this using the SUB instruction to subtract zero from the number, which effectively removes any negative sign.
\end{tcolorbox}

\begin{tcolorbox}[colback=white, colframe=black!20, coltitle=black, boxrule=0.5pt, fontupper=\small, title={\scriptsize\bfseries Submitted code}]
\begin{verbatim}
loop:
    INBOX
    JUMPZ end
    JUMPN skip_output
    OUTBOX
    JUMP loop
skip_output:
    SUB 0
    OUTBOX
    JUMP loop
end:
\end{verbatim}
\end{tcolorbox}

\textbf{Execution:} \texttt{Runtime error: At IP 6: Empty tile 0}

\begin{tcolorbox}[colback=white, colframe=black!20, coltitle=black, boxrule=0.5pt, fontupper=\small, title={\scriptsize\bfseries Judge verdict}]
\textbf{Plan 1:} The reasoning names the absolute-value task but applies an invalid arithmetic shortcut---``subtract zero to remove the negative sign'' misunderstands how SUB works in HRM. Subtracting an uninitialized tile does not compute the absolute value.

\textbf{Impl 1:} The program is syntactically valid and the control flow (branch on sign) is structurally reasonable, but it crashes at runtime because tile~0 was never initialized. This is a conceptual failure, not a formatting one.
\end{tcolorbox}
\end{tcolorbox}

\begin{tcolorbox}[breakable, colback=black!8, colframe=black!40, coltitle=black, title={\small\bfseries Example 3: SIC-1 --- First Assessment \hfill Plan: 2 \quad Impl: 1}, fontupper=\small, boxrule=0.8pt]
\textbf{Model:} Codestral 22B \quad \textbf{Category:} Correct plan, failed execution

\begin{tcolorbox}[colback=white, colframe=black!20, coltitle=black, boxrule=0.5pt, fontupper=\small, title={\scriptsize\bfseries Model reasoning}]
The task is to create a program that reads an input value and outputs it unchanged. Since we can't directly write to memory address 254 (@OUT), we need to manipulate the values in memory.
\end{tcolorbox}

\begin{tcolorbox}[colback=white, colframe=black!20, coltitle=black, boxrule=0.5pt, fontupper=\small, title={\scriptsize\bfseries Submitted code}]
\begin{verbatim}
@LOOP:
    subleq @TMP1, @TMP1
    subleq @TMP1, @INP
    subleq @RESULT, @RESULT
    subleq @RESULT, @TMP1
    subleq @TMP2, @TMP2
    subleq @INP, @IN
    subleq @TMP2, @INP
    subleq @RESULT, @TMP2
    subleq @OUT, @OUT
    subleq @NEG, @NEG
    subleq @NEG, @RESULT
    subleq @OUT, @NEG
    subleq @ZERO, @ZERO, @LOOP
\end{verbatim}
\end{tcolorbox}

\textbf{Execution:} \texttt{Runtime error: Cycle limit exceeded: 100000 cycles}

\begin{tcolorbox}[colback=white, colframe=black!20, coltitle=black, boxrule=0.5pt, fontupper=\small, title={\scriptsize\bfseries Judge verdict}]
\textbf{Plan 2:} The reasoning correctly identifies the identity transformation (read input, output unchanged) and recognizes that SIC-1 requires indirect memory manipulation since there is no direct copy instruction.

\textbf{Impl 1:} The program is syntactically valid and uses a plausible sequence of subtract-and-negate operations, but the control flow never reaches a successful terminating state. The loop runs indefinitely, exceeding the 100,000-cycle limit.
\end{tcolorbox}
\end{tcolorbox}

\begin{tcolorbox}[colback=black!8, colframe=black!40, coltitle=black, title={\small\bfseries Example 4: MHRD --- MUX \hfill Plan: 1 \quad Impl: 1}, fontupper=\small, boxrule=0.8pt]
\textbf{Model:} Codestral 22B \quad \textbf{Category:} Partial plan, partial execution

\begin{tcolorbox}[colback=white, colframe=black!20, coltitle=black, boxrule=0.5pt, fontupper=\small, title={\scriptsize\bfseries Model reasoning}]
A MUX selects one of two inputs based on a selector. Use AND to multiply in1 with the inverse of sel. Use AND to multiply in2 with sel. Then use OR to combine the results.
\end{tcolorbox}

\begin{tcolorbox}[colback=white, colframe=black!20, coltitle=black, boxrule=0.5pt, fontupper=\small, title={\scriptsize\bfseries Submitted code}]
\begin{verbatim}
Inputs: in1, in2, sel;
Outputs: out;
Parts: n1 NAND, n2 NAND,
  n3 NAND, n4 NAND, n5 NAND;
Wires:
  sel->n1.in1, sel->n1.in2,
  in1->n2.in1, n1.out->n2.in2,
  sel->n3.in1, sel->n3.in2,
  in2->n4.in1, n3.out->n4.in2,
  n2.out->n5.in1,
  n4.out->n5.in2,
  n5.out->out;
\end{verbatim}
\end{tcolorbox}

\textbf{Execution:} Passed 5 of 8 truth-table entries (reward: 0.44)

\begin{tcolorbox}[colback=white, colframe=black!20, coltitle=black, boxrule=0.5pt, fontupper=\small, title={\scriptsize\bfseries Judge verdict}]
\textbf{Plan 1:} The reasoning states the textbook MUX formula ($\text{out} = (\text{in1} \wedge \overline{\text{sel}}) \vee (\text{in2} \wedge \text{sel})$) but does not correctly decompose it into NAND-only logic---the description conflates AND with NAND.

\textbf{Impl 1:} The circuit parses and executes but gate n3 duplicates the NOT of sel instead of passing sel through, so 3 of 8 truth-table rows produce wrong output.
\end{tcolorbox}
\end{tcolorbox}

\begin{tcolorbox}[colback=black!8, colframe=black!40, coltitle=black, title={\small\bfseries Example 5: SIC-1 --- Sequence Sum \hfill Plan: 1 \quad Impl: 0}, fontupper=\small, boxrule=0.8pt]
\textbf{Model:} Codestral 22B \quad \textbf{Category:} Syntax/format failure

\begin{tcolorbox}[colback=white, colframe=black!20, coltitle=black, boxrule=0.5pt, fontupper=\small, title={\scriptsize\bfseries Model reasoning}]
I need to continuously read values from input and add them. The sequence is terminated by zero. I can use a loop that reads values, adds them to a running sum, and checks if the input was zero.
\end{tcolorbox}

\begin{tcolorbox}[colback=white, colframe=black!20, coltitle=black, boxrule=0.5pt, fontupper=\small, title={\scriptsize\bfseries Submitted code}]
\begin{verbatim}
@loop:
  ; Read from input
  subleq @IN, @VAL
  ; Add to sum
  subleq @SUM, @VAL
  ; Check if zero
  subleq @TEMP, @VAL
  subleq @END, @TEMP
  subleq @ZERO, @ZERO, @loop
@END:
  subleq @NEG_SUM, @SUM
  ...
.data 0
@IN .data 253
@OUT .data 254
@VAL .data 0
@TEMP .data 0
@ZERO .data 0
@SUM .data 0
@NEG_SUM .data 0
@HALT .data 255
\end{verbatim}
\end{tcolorbox}

\textbf{Execution:} \texttt{Unknown command: "@IN"}

\begin{tcolorbox}[colback=white, colframe=black!20, coltitle=black, boxrule=0.5pt, fontupper=\small, title={\scriptsize\bfseries Judge verdict}]
\textbf{Plan 1:} The reasoning captures the sentinel-sum structure (loop, accumulate, check for zero terminator) but does not address how to implement addition using only subtract-and-branch.

\textbf{Impl 0:} The code uses \texttt{@IN} as a standalone line, which the SIC-1 assembler treats as an unknown command. The assembler only recognizes \texttt{subleq} and \texttt{.data} as valid statements. The label-as-command confusion prevents parsing.
\end{tcolorbox}
\end{tcolorbox}

\begin{tcolorbox}[colback=black!8, colframe=black!40, coltitle=black, title={\small\bfseries Example 6: BOX-256 --- Corner Squares \hfill Plan: 1 \quad Impl: 1}, fontupper=\small, boxrule=0.8pt]
\textbf{Model:} Codestral 22B \quad \textbf{Category:} Partial plan, partial execution

\begin{tcolorbox}[colback=white, colframe=black!20, coltitle=black, boxrule=0.5pt, fontupper=\small, title={\scriptsize\bfseries Model reasoning}]
This problem can be solved by drawing each corner square individually. Since we are only required to draw hollow squares, this can be done by setting pixels at the border of each square.
\end{tcolorbox}

\begin{tcolorbox}[colback=white, colframe=black!20, coltitle=black, boxrule=0.5pt, fontupper=\small, title={\scriptsize\bfseries Submitted code}]
\begin{verbatim}
; top-left square (0,0)-(7,7)
MOV #1 @0 #8
MOV #1 @15 @8
ADD #64 #127 @255
MOV #1 @136 #8
ADD #15 #143 @255
MOV #1 @255 #8
; top-right square (8,0)-(15,7)
ADD #16 #15 @254
MOV #1 @254 #8
\end{verbatim}
\end{tcolorbox}

\textbf{Execution:} 0 of 1 tests passed (reward: 0.40)

\begin{tcolorbox}[colback=white, colframe=black!20, coltitle=black, boxrule=0.5pt, fontupper=\small, title={\scriptsize\bfseries Judge verdict}]
\textbf{Plan 1:} The reasoning correctly identifies that the target consists of hollow square borders at the corners and proposes drawing each individually. However, it uses 8$\times$8 regions instead of the benchmark's 4$\times$4 squares, and does not account for the specific corner colors.

\textbf{Impl 1:} The program parses and executes, producing a partial pixel pattern (reward: 0.40). The border-drawing approach is structurally sound but targets the wrong coordinates and sizes, so the output does not match the target image.
\end{tcolorbox}
\end{tcolorbox}

\begin{tcolorbox}[colback=black!8, colframe=black!40, coltitle=black, title={\small\bfseries Example 7: MHRD --- DEMUX \hfill Plan: 1 \quad Impl: 1}, fontupper=\small, boxrule=0.8pt]
\textbf{Model:} Phi-4-Reasoning \quad \textbf{Category:} Wrong plan

\begin{tcolorbox}[colback=white, colframe=black!20, coltitle=black, boxrule=0.5pt, fontupper=\small, title={\scriptsize\bfseries Model reasoning}]
DEMUX means a demultiplexer. We have input signals: in, sel. Route input to out1 when sel=0, out2 when sel=1. If sel=0 then out1 gets the input and out2 is 0.
\end{tcolorbox}

\begin{tcolorbox}[colback=white, colframe=black!20, coltitle=black, boxrule=0.5pt, fontupper=\small, title={\scriptsize\bfseries Submitted code}]
\begin{verbatim}
Inputs: in;
Outputs: out;
Parts: n NAND;
Wires:
   in -> n.in1, in -> n.in2,
   n.out -> out;
\end{verbatim}
\end{tcolorbox}

\textbf{Execution:} Passed 0 of 4 tests

\begin{tcolorbox}[colback=white, colframe=black!20, coltitle=black, boxrule=0.5pt, fontupper=\small, title={\scriptsize\bfseries Judge verdict}]
\textbf{Plan 1:} The reasoning correctly describes the DEMUX interface (route input to one of two outputs based on selector), but the submitted netlist declares only one input and one output---it implements a NOT gate, not a demultiplexer.

\textbf{Impl 1:} The circuit parses and executes as a valid MHRD program, but it has the wrong interface (missing \texttt{sel}, \texttt{out1}, \texttt{out2}) and implements the wrong function entirely. All 4 truth-table entries fail.
\end{tcolorbox}
\end{tcolorbox}

\end{document}

%% file: figures/fig_overview.tex
\begin{figure*}[t!]
\centering
\includegraphics[width=\textwidth]{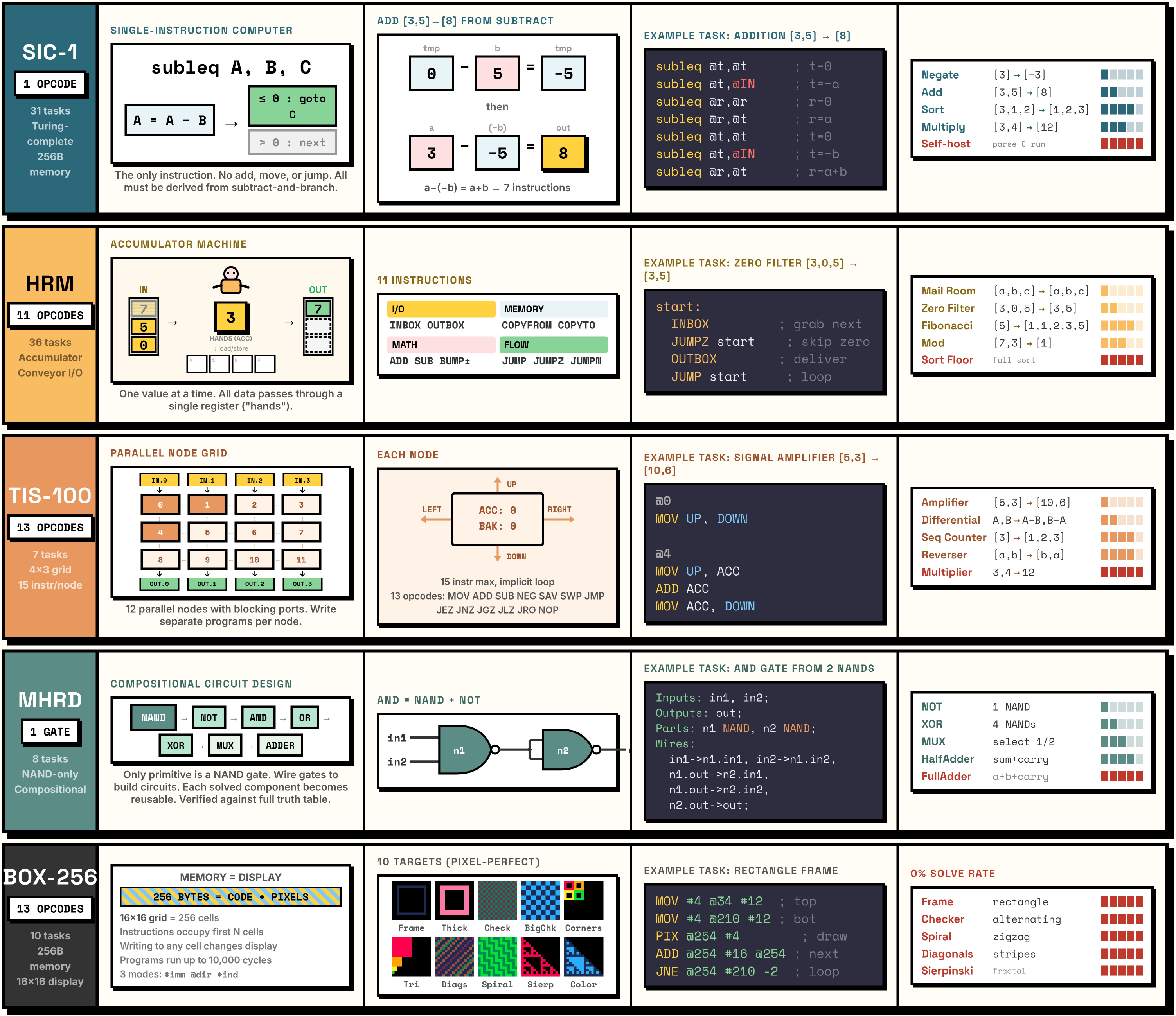}
\caption{Overview of the five ISA-Bench environments. Each panel shows the computational model, an example task or code snippet, and the task range for one game. The environments progress from a single-instruction machine (\textbf{SIC-1}, where even addition requires a multi-step derivation from subtract-and-branch) through sequential architectures with increasing structure (\textbf{HRM}'s accumulator with conveyor I/O; \textbf{TIS-100}'s parallel node grid with blocking inter-node communication) to declarative design (\textbf{MHRD}'s compositional circuit wiring from a single NAND primitive) and spatial output generation (\textbf{BOX-256}'s memory-mapped pixel display). This progression tests computational primitives: algebraic derivation, sequential planning, parallel coordination, structural composition, and spatial reasoning.}
\label{fig:overview}
\end{figure*}

%% file: figures/evaluation_pipeline.tex
\begin{figure}[t!]
\centering
\includegraphics[width=0.75\columnwidth]{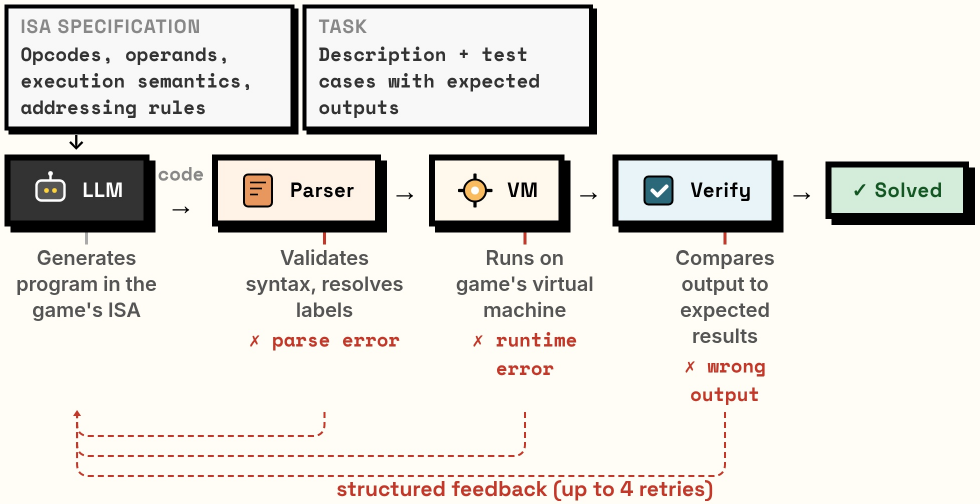}
\caption{Evaluation pipeline for each task. The model receives an ISA specification and task description, then generates a program that passes through three stages: parsing (syntax validation), execution (running on the game's VM), and verification (comparing outputs to test cases). Each stage can produce a distinct failure type. On failure, structured feedback is returned for up to four retries (five attempts in total).}
\label{fig:pipeline}
\end{figure}

%% file: figures/fig_solverate.tex
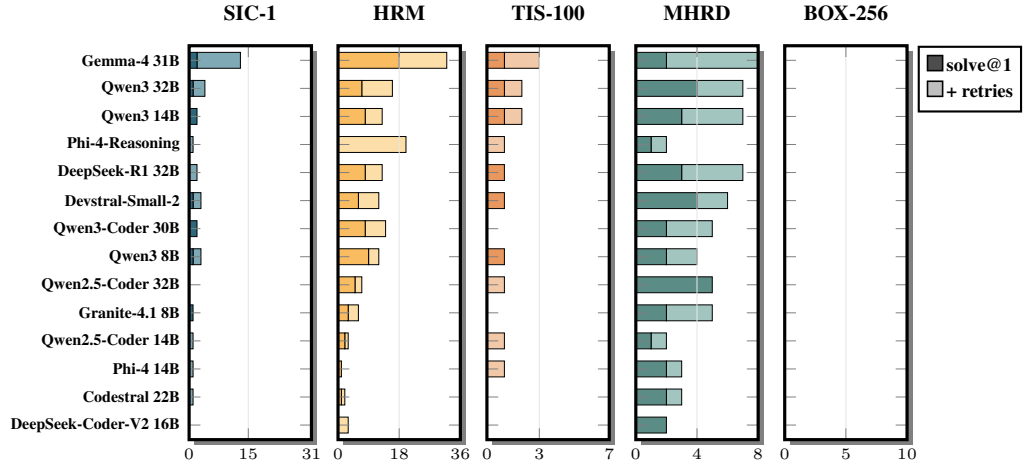
\begin{figure*}[t!]
\centering
\pgfplotsset{
    neobrutbar/.style={
        xbar stacked,
        bar width=6pt,
        height=6.8cm,
        width=3.2cm,
        enlarge y limits=0.04,
        xmin=0,
        xmajorgrids=true,
        grid style={gray!20, line width=0.4pt},
        tick label style={font=\tiny\bfseries},
        title style={font=\footnotesize\bfseries, at={(0.5,1.0)}, anchor=south},
        ytick=data,
        y dir=reverse,
        every axis plot/.append style={draw=black, line width=0.3pt},
        axis on top,
    },
}

\begin{tikzpicture}
\begin{groupplot}[
    group style={
        group size=5 by 1,
        horizontal sep=10pt,
        y descriptions at=edge left,
    },
    neobrutbar,
    symbolic y coords={
        Gemma-4 31B,
        Qwen3 32B,
        Qwen3 14B,
        Phi-4-Reasoning,
        DeepSeek-R1 32B,
        Devstral-Small-2,
        Qwen3-Coder 30B,
        Qwen3 8B,
        Qwen2.5-Coder 32B,
        Granite-4.1 8B,
        Qwen2.5-Coder 14B,
        Phi-4 14B,
        Codestral 22B,
        DeepSeek-Coder-V2 16B,
    },
]

\nextgroupplot[title={SIC-1}, xmax=31, xtick={0,15,31},
    axis background/.style={fill=white},
    axis line style={line width=1.2pt},
    yticklabels={
        Gemma-4 31B,
        Qwen3 32B,
        Qwen3 14B,
        Phi-4-Reasoning,
        DeepSeek-R1 32B,
        Devstral-Small-2,
        Qwen3-Coder 30B,
        Qwen3 8B,
        Qwen2.5-Coder 32B,
        Granite-4.1 8B,
        Qwen2.5-Coder 14B,
        Phi-4 14B,
        Codestral 22B,
        DeepSeek-Coder-V2 16B,
    },
    yticklabel style={font=\tiny\bfseries, anchor=east},
]
\addplot[fill=clrSIC] coordinates {
    (2,Gemma-4 31B) (1,Qwen3 32B) (2,Qwen3 14B) (0,Phi-4-Reasoning)
    (0,DeepSeek-R1 32B) (1,Devstral-Small-2) (2,Qwen3-Coder 30B) (1,Qwen3 8B)
    (0,Qwen2.5-Coder 32B) (1,Granite-4.1 8B) (0,Qwen2.5-Coder 14B)
    (0,Phi-4 14B) (0,Codestral 22B) (0,DeepSeek-Coder-V2 16B)
};
\addplot[fill=clrSIClt] coordinates {
    (11,Gemma-4 31B) (3,Qwen3 32B) (0,Qwen3 14B) (1,Phi-4-Reasoning)
    (2,DeepSeek-R1 32B) (2,Devstral-Small-2) (0,Qwen3-Coder 30B) (2,Qwen3 8B)
    (0,Qwen2.5-Coder 32B) (0,Granite-4.1 8B) (1,Qwen2.5-Coder 14B)
    (1,Phi-4 14B) (1,Codestral 22B) (0,DeepSeek-Coder-V2 16B)
};

\nextgroupplot[title={HRM}, xmax=36, xtick={0,18,36}, yticklabels={},
    axis background/.style={fill=white}, axis line style={line width=1.2pt}]
\addplot[fill=clrHRM] coordinates {
    (18,Gemma-4 31B) (7,Qwen3 32B) (8,Qwen3 14B) (0,Phi-4-Reasoning)
    (8,DeepSeek-R1 32B) (6,Devstral-Small-2) (8,Qwen3-Coder 30B) (9,Qwen3 8B)
    (5,Qwen2.5-Coder 32B) (3,Granite-4.1 8B) (2,Qwen2.5-Coder 14B)
    (1,Phi-4 14B) (1,Codestral 22B) (0,DeepSeek-Coder-V2 16B)
};
\addplot[fill=clrHRMlt] coordinates {
    (14,Gemma-4 31B) (9,Qwen3 32B) (5,Qwen3 14B) (20,Phi-4-Reasoning)
    (5,DeepSeek-R1 32B) (6,Devstral-Small-2) (6,Qwen3-Coder 30B) (3,Qwen3 8B)
    (2,Qwen2.5-Coder 32B) (3,Granite-4.1 8B) (1,Qwen2.5-Coder 14B)
    (0,Phi-4 14B) (1,Codestral 22B) (3,DeepSeek-Coder-V2 16B)
};

\nextgroupplot[title={TIS-100}, xmax=7, xtick={0,3,7}, yticklabels={},
    axis background/.style={fill=white}, axis line style={line width=1.2pt}]
\addplot[fill=clrTIS] coordinates {
    (1,Gemma-4 31B) (1,Qwen3 32B) (1,Qwen3 14B) (0,Phi-4-Reasoning)
    (1,DeepSeek-R1 32B) (1,Devstral-Small-2) (0,Qwen3-Coder 30B) (1,Qwen3 8B)
    (0,Qwen2.5-Coder 32B) (0,Granite-4.1 8B) (0,Qwen2.5-Coder 14B)
    (0,Phi-4 14B) (0,Codestral 22B) (0,DeepSeek-Coder-V2 16B)
};
\addplot[fill=clrTISlt] coordinates {
    (2,Gemma-4 31B) (1,Qwen3 32B) (1,Qwen3 14B) (1,Phi-4-Reasoning)
    (0,DeepSeek-R1 32B) (0,Devstral-Small-2) (0,Qwen3-Coder 30B) (0,Qwen3 8B)
    (1,Qwen2.5-Coder 32B) (0,Granite-4.1 8B) (1,Qwen2.5-Coder 14B)
    (1,Phi-4 14B) (0,Codestral 22B) (0,DeepSeek-Coder-V2 16B)
};

\nextgroupplot[title={MHRD}, xmax=8, xtick={0,4,8}, yticklabels={},
    axis background/.style={fill=white}, axis line style={line width=1.2pt}]
\addplot[fill=clrMHRD] coordinates {
    (2,Gemma-4 31B) (4,Qwen3 32B) (3,Qwen3 14B) (1,Phi-4-Reasoning)
    (3,DeepSeek-R1 32B) (4,Devstral-Small-2) (2,Qwen3-Coder 30B) (2,Qwen3 8B)
    (5,Qwen2.5-Coder 32B) (2,Granite-4.1 8B) (1,Qwen2.5-Coder 14B)
    (2,Phi-4 14B) (2,Codestral 22B) (2,DeepSeek-Coder-V2 16B)
};
\addplot[fill=clrMHRDlt] coordinates {
    (6,Gemma-4 31B) (3,Qwen3 32B) (4,Qwen3 14B) (1,Phi-4-Reasoning)
    (4,DeepSeek-R1 32B) (2,Devstral-Small-2) (3,Qwen3-Coder 30B) (2,Qwen3 8B)
    (0,Qwen2.5-Coder 32B) (3,Granite-4.1 8B) (1,Qwen2.5-Coder 14B)
    (1,Phi-4 14B) (1,Codestral 22B) (0,DeepSeek-Coder-V2 16B)
};

\nextgroupplot[title={BOX-256}, xmax=10, xtick={0,5,10}, yticklabels={},
    axis background/.style={fill=white}, axis line style={line width=1.2pt}]
\addplot[fill=clrBOX] coordinates {
    (0,Gemma-4 31B) (0,Qwen3 32B) (0,Qwen3 14B) (0,Phi-4-Reasoning)
    (0,DeepSeek-R1 32B) (0,Devstral-Small-2) (0,Qwen3-Coder 30B) (0,Qwen3 8B)
    (0,Qwen2.5-Coder 32B) (0,Granite-4.1 8B) (0,Qwen2.5-Coder 14B)
    (0,Phi-4 14B) (0,Codestral 22B) (0,DeepSeek-Coder-V2 16B)
};
\addplot[fill=clrBOXlt] coordinates {
    (0,Gemma-4 31B) (0,Qwen3 32B) (0,Qwen3 14B) (0,Phi-4-Reasoning)
    (0,DeepSeek-R1 32B) (0,Devstral-Small-2) (0,Qwen3-Coder 30B) (0,Qwen3 8B)
    (0,Qwen2.5-Coder 32B) (0,Granite-4.1 8B) (0,Qwen2.5-Coder 14B)
    (0,Phi-4 14B) (0,Codestral 22B) (0,DeepSeek-Coder-V2 16B)
};

\end{groupplot}
\begin{scope}[on background layer]
\fill[black!50] ([xshift=2pt,yshift=-2pt]group c1r1.south west) rectangle ([xshift=2pt,yshift=-2pt]group c1r1.north east);
\fill[black!50] ([xshift=2pt,yshift=-2pt]group c2r1.south west) rectangle ([xshift=2pt,yshift=-2pt]group c2r1.north east);
\fill[black!50] ([xshift=2pt,yshift=-2pt]group c3r1.south west) rectangle ([xshift=2pt,yshift=-2pt]group c3r1.north east);
\fill[black!50] ([xshift=2pt,yshift=-2pt]group c4r1.south west) rectangle ([xshift=2pt,yshift=-2pt]group c4r1.north east);
\fill[black!50] ([xshift=2pt,yshift=-2pt]group c5r1.south west) rectangle ([xshift=2pt,yshift=-2pt]group c5r1.north east);
\end{scope}
\node[anchor=north west, font=\scriptsize\bfseries, fill=white, draw=black, line width=0.8pt, inner sep=3pt] at ([xshift=4pt]group c5r1.north east) {%
    \begin{tabular}{@{}l@{}}
    \tikz{\fill[black!70,draw=black,line width=0.6pt] (0,0) rectangle (0.2,0.2);}\,solve@1 \\[1pt]
    \tikz{\fill[black!25,draw=black,line width=0.6pt] (0,0) rectangle (0.2,0.2);}\,+ retries
    \end{tabular}
};
\end{tikzpicture}
\caption{Tasks solved per game for 14 models, in the same order as Table~\ref{tab:model_results}. Dark bars show solve@1; light extensions show additional solves obtained through retries, so the full stacked bar shows solve@5.}
\label{fig:solve_rates}
\end{figure*}

%% file: figures/fig_failure.tex
\begin{figure*}[t!]
\centering
\newcommand{\failurepie}[7]{%
\begin{scope}[xshift=#2cm]
\pgfmathsetmacro{\failuremass}{1-#4/#3}
\pgfmathsetmacro{\parseend}{90-360*\failuremass*(#5/(#5+#6+#7))}
\pgfmathsetmacro{\runtimeend}{\parseend-360*\failuremass*(#6/(#5+#6+#7))}
\pgfmathsetmacro{\wrongend}{90-360*\failuremass}
\node[font=\scriptsize\bfseries] at (0,1.35) {#1};
\fill[clrParse] (0,0) -- (90:1) arc (90:\parseend:1) -- cycle;
\ifnum#6>0
\fill[clrRuntime] (0,0) -- (\parseend:1) arc (\parseend:\runtimeend:1) -- cycle;
\fi
\fill[clrWrong] (0,0) -- (\runtimeend:1) arc (\runtimeend:\wrongend:1) -- cycle;
\ifnum#4>0
\fill[clrSolved] (0,0) -- (\wrongend:1) arc (\wrongend:-270:1) -- cycle;
\fi
\draw[black, line width=1pt] (0,0) circle (1);
\foreach \a in {90,\parseend,\runtimeend,\wrongend} {
    \draw[black, line width=0.3pt] (0,0) -- (\a:1);
}
\end{scope}%
}
\resizebox{\textwidth}{!}{%
\begin{tikzpicture}
\input{figures/failure_data}

\begin{scope}[on background layer]
\foreach \x in {0,3,6,9,12} {
    \fill[black!40] (\x+0.07,-0.07) circle (1);
}
\end{scope}

\node[anchor=north, font=\scriptsize\bfseries] at (6, -1.5) {%
    \tikz{\fill[clrParse,draw=black,line width=0.5pt] (0,0) rectangle (0.2,0.2);}\,Parse error \quad
    \tikz{\fill[clrRuntime,draw=black,line width=0.5pt] (0,0) rectangle (0.2,0.2);}\,Runtime error \quad
    \tikz{\fill[clrWrong,draw=black,line width=0.5pt] (0,0) rectangle (0.2,0.2);}\,Wrong output \quad
    \tikz{\fill[clrSolved,draw=black,line width=0.5pt] (0,0) rectangle (0.2,0.2);}\,Solved
};
\end{tikzpicture}
}
\caption{First-attempt outcome distributions by game, aggregated across 14 models.}
\label{fig:failure_breakdown}
\end{figure*}
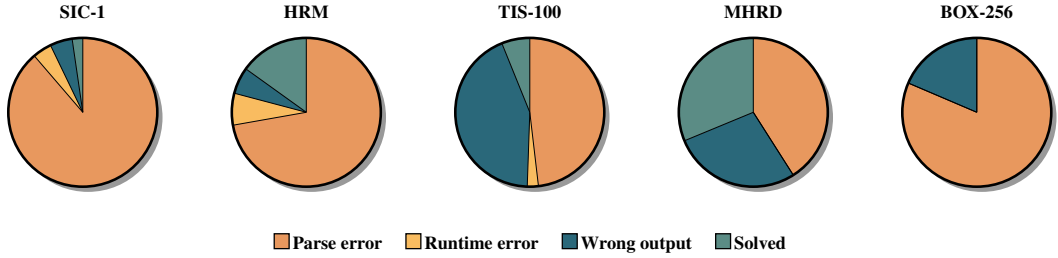

%% file: figures/failure_data.tex
\failurepie{SIC-1}{0}{434}{10}{385}{18}{21}
\failurepie{HRM}{3}{504}{76}{364}{35}{29}
\failurepie{TIS-100}{6}{98}{6}{47}{2}{43}
\failurepie{MHRD}{9}{112}{35}{46}{0}{31}
\failurepie{BOX-256}{12}{140}{0}{114}{0}{26}

%% file: figures/fig_box256targets.tex
\begin{figure*}[t]
\centering
\includegraphics[width=\textwidth]{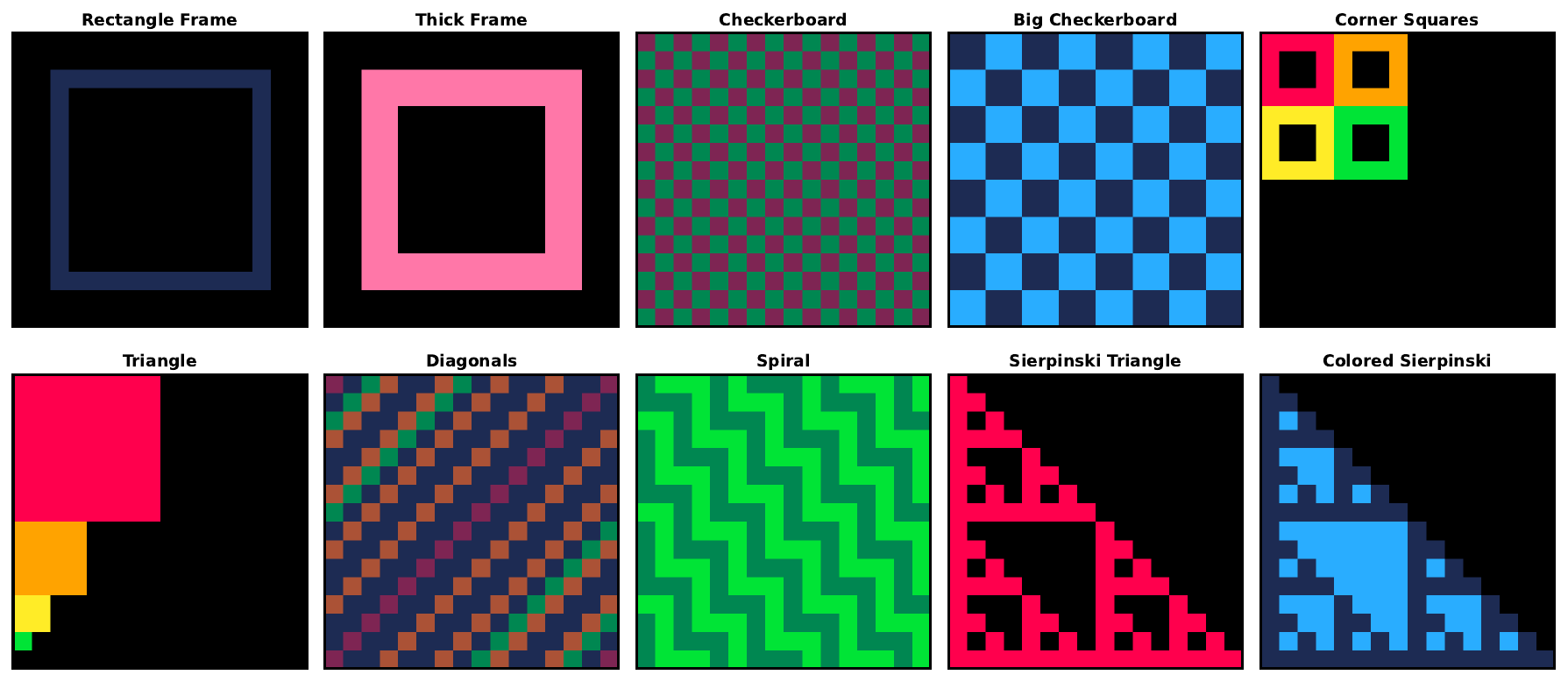}
\caption{Target pixel patterns for all 10 BOX-256 tasks. Each grid is $16\times16$ pixels. The program must produce the exact pattern by writing color values to shared code/display memory.}
\label{fig:box256_targets}
\end{figure*}

%% file: appendix_prompts.tex
\begin{tcolorbox}[colback=clrSIC!10, colframe=clrSIC, coltitle=white, title={\small\bfseries System prompt: SIC-1}, boxrule=0.8pt, breakable, left=2pt, right=2pt, top=2pt, bottom=2pt]
\begin{Verbatim}[breaklines, breakanywhere, fontsize=\scriptsize, commandchars=none]
You are an expert SIC-1 programmer. Your task is to write subleq programs that transform input sequences into specified output sequences.

# SIC-1 SPECIFICATION

## Architecture
SIC-1 is a Single Instruction Computer. The only instruction is `subleq` (subtract and branch if less than or equal to zero).

- **Memory**: 256 bytes (addresses 0-255), each byte is signed (-128 to 127)
- **Execution**: Programs start at address 0, each instruction is 3 bytes
- **I/O**: Special memory addresses for input/output

## The SUBLEQ Instruction

```
subleq A, B [, C]
```

**Semantics**:
1. `mem[A] = mem[A] - mem[B]`
2. If `mem[A] <= 0`: jump to address C
3. Else: continue to next instruction (PC += 3)

When C is omitted, it defaults to the next instruction (no branch).

## Special Addresses
| Address | Label | Behavior |
|---------|-------|----------|
| 253 | `@IN` | Reading returns next input value |
| 254 | `@OUT` | Writing outputs the value |
| 255 | `@HALT` | Any access halts execution |

## Assembler Syntax
```
@label:              ; define a label
subleq @a, @b, @c    ; subtract and branch
.data 42             ; data value
.data 'A'            ; character (ASCII)
@label+1             ; label with offset
-@label              ; negated address value
```

## Key Patterns

### Pattern 1: Output = 0 - Input (negate)
```
subleq @OUT, @IN     ; @OUT = @OUT - @IN = 0 - input = -input
```

### Pattern 2: Unconditional Jump
```
subleq @zero, @zero, @target   ; 0-0=0, always <=0, so always jumps
@zero: .data 0
```

### Pattern 3: Zero a Memory Location
```
subleq @x, @x        ; x = x - x = 0
```

### Pattern 4: Copy Value (A to B)
```
subleq @B, @B        ; B = 0
subleq @tmp, @tmp    ; tmp = 0
subleq @tmp, @A      ; tmp = 0 - A = -A
subleq @B, @tmp      ; B = 0 - (-A) = A
```

### Pattern 5: Add Two Values
```
; Result = A + B
subleq @result, @result  ; result = 0
subleq @tmp, @tmp        ; tmp = 0
subleq @tmp, @A          ; tmp = -A
subleq @result, @tmp     ; result = A
subleq @tmp, @tmp        ; tmp = 0
subleq @tmp, @B          ; tmp = -B
subleq @result, @tmp     ; result = A + B
```

# RESPONSE FORMAT

Reason through the problem in <think> tags, then provide your solution in <answer> tags.
The code inside <answer> will be executed directly.

Remember: subleq is counterintuitive. Think carefully about signs and the subtract-and-branch semantics.
\end{Verbatim}
\end{tcolorbox}
\vspace{4pt}

\begin{tcolorbox}[colback=clrHRM!15, colframe=clrHRM!80!black, coltitle=white, title={\small\bfseries System prompt: HRM}, boxrule=0.8pt, breakable, left=2pt, right=2pt, top=2pt, bottom=2pt]
\begin{Verbatim}[breaklines, breakanywhere, fontsize=\scriptsize, commandchars=none]
You are an expert Human Resource Machine (HRM) programmer. Your task is to write programs that transform input sequences into specified output sequences.

# HRM SPECIFICATION

## Architecture
- **Hands (Accumulator)**: Holds exactly one value. Operations require data in hands.
- **Floor Tiles (Memory)**: Numbered slots (0, 1, 2, ...) for storing values.
- **INBOX**: Input conveyor. Reading when empty terminates the program.
- **OUTBOX**: Output conveyor. Program must produce exact expected sequence.
- **Values**: Integers (-999 to 999) or single uppercase letters (A-Z).

## Instruction Set

### Input/Output
| Instruction | Effect |
|-------------|--------|
| `INBOX` | Take next value from INBOX into hands. Empty INBOX terminates program. |
| `OUTBOX` | Put value from hands onto OUTBOX. Hands become empty. |

### Memory Access
| Instruction | Effect |
|-------------|--------|
| `COPYFROM n` | Copy value from tile n into hands. |
| `COPYFROM [n]` | Indirect: tile n contains address, copy from that address. |
| `COPYTO n` | Copy value from hands to tile n. Hands keep the value. |
| `COPYTO [n]` | Indirect addressing for destination. |

### Arithmetic
| Instruction | Effect |
|-------------|--------|
| `ADD n` | hands = hands + tile[n]. Works with [n] for indirect. |
| `SUB n` | hands = hands - tile[n]. Works with [n] for indirect. |
| `BUMPUP n` | Increment tile[n] by 1, copy result to hands. |
| `BUMPDN n` | Decrement tile[n] by 1, copy result to hands. |

### Control Flow
| Instruction | Effect |
|-------------|--------|
| `JUMP label` | Unconditional jump to label. |
| `JUMPZ label` | Jump if hands == 0. |
| `JUMPN label` | Jump if hands < 0. |

## Syntax
Labels are defined with a trailing colon on their own line:
```
myLabel:
    INBOX
    OUTBOX
    JUMP myLabel
```

# RESPONSE FORMAT

Reason through the problem in <think> tags, then provide your solution in <answer> tags.
The code inside <answer> will be executed directly.
\end{Verbatim}
\end{tcolorbox}
\vspace{4pt}

\begin{tcolorbox}[colback=clrTIS!10, colframe=clrTIS, coltitle=white, title={\small\bfseries System prompt: TIS-100}, boxrule=0.8pt, breakable, left=2pt, right=2pt, top=2pt, bottom=2pt]
\begin{Verbatim}[breaklines, breakanywhere, fontsize=\scriptsize, commandchars=none]
You are an expert TIS-100 programmer. Your task is to write programs that transform input streams into output streams using a grid of parallel compute nodes.

# TIS-100 SPECIFICATION

## Architecture
- **Grid**: 4 columns x 3 rows = 12 compute nodes
- **Node numbering**: 0-3 (top), 4-7 (middle), 8-11 (bottom)
- **Registers**: ACC (accumulator), BAK (backup, not directly accessible)
- **Value range**: -999 to 999
- **I/O**: Inputs feed from above row 0, outputs collect below row 2

## Grid Layout
```
  [IN.0] [IN.1] [IN.2] [IN.3]   <- Input streams (columns 0-3)
    v      v      v      v
  [ 0 ]--[ 1 ]--[ 2 ]--[ 3 ]   <- Row 0 (nodes 0-3)
    v      v      v      v
  [ 4 ]--[ 5 ]--[ 6 ]--[ 7 ]   <- Row 1 (nodes 4-7)
    v      v      v      v
  [ 8 ]--[ 9 ]--[10 ]--[11 ]   <- Row 2 (nodes 8-11)
    v      v      v      v
 [OUT.0][OUT.1][OUT.2][OUT.3]  <- Output streams (columns 0-3)
```

## Instruction Set

### Data Movement
| Instruction | Effect |
|-------------|--------|
| `MOV SRC, DST` | Copy value from SRC to DST |

**SRC**: ACC, NIL, port (UP/DOWN/LEFT/RIGHT/ANY/LAST), immediate (-999 to 999)
**DST**: ACC, NIL, port (UP/DOWN/LEFT/RIGHT/ANY/LAST)

### Arithmetic
| Instruction | Effect |
|-------------|--------|
| `ADD SRC` | ACC = ACC + SRC |
| `SUB SRC` | ACC = ACC - SRC |
| `NEG` | ACC = -ACC |

### Register Operations
| Instruction | Effect |
|-------------|--------|
| `SAV` | BAK = ACC (save to backup) |
| `SWP` | Swap ACC and BAK |

### Control Flow
| Instruction | Effect |
|-------------|--------|
| `JMP label` | Unconditional jump |
| `JEZ label` | Jump if ACC == 0 |
| `JNZ label` | Jump if ACC != 0 |
| `JGZ label` | Jump if ACC > 0 |
| `JLZ label` | Jump if ACC < 0 |
| `JRO SRC` | Jump relative by SRC instructions |

## Port Communication
Ports are **blocking**: a node waits until both sender and receiver are ready.
- `UP`, `DOWN`, `LEFT`, `RIGHT`: Cardinal directions
- `ANY`: Read from first available port
- `LAST`: Use the port that ANY last used

## Program Execution
- Each node runs its program **independently and in parallel** with other nodes.
- Programs **implicitly loop**: after the last instruction, execution wraps back to the first instruction.
- Each node has a **15 instruction limit**. Programs longer than 15 lines will be rejected.
- A node with no code does nothing (acts as a wall to port communication).

## Program Syntax
Use `@N` to specify which node runs each code block:
```
@0
MOV UP, ACC
ADD ACC
MOV ACC, DOWN

@4
MOV UP, DOWN
```

# RESPONSE FORMAT

Reason through the problem in <think> tags, then provide your solution in <answer> tags.
The code inside <answer> will be executed directly.

Key insight: Trace the data flow from input columns through the grid to output columns. Every node in the path needs code to forward data.
\end{Verbatim}
\end{tcolorbox}
\vspace{4pt}

\begin{tcolorbox}[colback=clrMHRD!10, colframe=clrMHRD!80!black, coltitle=white, title={\small\bfseries System prompt: MHRD}, boxrule=0.8pt, breakable, left=2pt, right=2pt, top=2pt, bottom=2pt]
\begin{Verbatim}[breaklines, breakanywhere, fontsize=\scriptsize, commandchars=none]
You are an expert MHRD hardware designer. Your task is to write HDL code to implement digital logic circuits using NAND gates.

# MHRD SPECIFICATION

## Overview
MHRD is a hardware design game where you build digital circuits from NAND gates. Each puzzle gives you a truth table to implement.

## HDL Syntax (4 Keywords)

```
Inputs: <input_list>;
Outputs: <output_list>;
Parts: <part_list>;
Wires: <wire_list>;
```

### Inputs/Outputs
```
Inputs: a, b;           // two single-bit inputs
Inputs: in[4];          // 4-bit bus: in[1], in[2], in[3], in[4]
Outputs: out;           // single-bit output
Outputs: sum, carry;    // multiple outputs
```

### Parts (Component Instances)
```
Parts: n1 NAND, n2 NAND;        // two NAND gates
Parts: inv NOT, a1 AND;          // using built components
```

### Wires (Connections)
```
Wires:
  a -> n1.in1,           // input a to n1's in1 port
  b -> n1.in2,           // input b to n1's in2 port
  n1.out -> out;         // n1's output to circuit output
```

## NAND Gate (The Only Primitive)

**Ports**: in1, in2, out

**Truth Table**:
| in1 | in2 | out |
|-----|-----|-----|
| 0 | 0 | 1 |
| 0 | 1 | 1 |
| 1 | 0 | 1 |
| 1 | 1 | 0 |

NAND outputs 0 only when BOTH inputs are 1.

## Building Blocks from NAND

### NOT (1 NAND): Connect input to both NAND inputs
```
NOT: in -> n.in1, in -> n.in2, n.out -> out
```

### AND (2 NANDs): NAND then NOT
```
AND: NAND(a,b) -> NOT -> out
```

### OR (3 NANDs): NOT both inputs, then NAND
```
OR: NOT(a) -> n3.in1, NOT(b) -> n3.in2, n3.out -> out
```

### XOR (4 NANDs): 
```
XOR: n1=NAND(a,b), n2=NAND(a,n1), n3=NAND(n1,b), n4=NAND(n2,n3)
```

# RESPONSE FORMAT

Reason through the problem in <think> tags, then provide your solution in <answer> tags.
The code inside <answer> will be executed directly.
\end{Verbatim}
\end{tcolorbox}
\vspace{4pt}

\begin{tcolorbox}[colback=clrBOX!15, colframe=black!60, coltitle=white, title={\small\bfseries System prompt: BOX-256}, boxrule=0.8pt, breakable, left=2pt, right=2pt, top=2pt, bottom=2pt]
\begin{Verbatim}[breaklines, breakanywhere, fontsize=\scriptsize, commandchars=none]
You are an expert BOX-256 assembly programmer. Your task is to write programs that draw specific pixel patterns on a 16x16 display.

# BOX-256 SPECIFICATION

## Architecture
- **Memory**: 256 bytes (addresses 0-255), shared between code and display
- **Display**: 16x16 grid where address = y*16 + x
- **Colors**: Values 0-15 (0=black). Only the low 4 bits matter for color.
- **All arithmetic is mod 256**: values wrap (255 + 1 = 0, 0 - 1 = 255)
- **Code lives in memory**: Instructions are 4 bytes each, loaded starting at address 0. Your code bytes ARE visible as pixels.
- **Execution**: Programs run for up to 10,000 cycles. Each cycle executes all threads.
- **Program wrapping**: After the last instruction, execution wraps to instruction 0.

## Addressing Modes
| Syntax | Name | Meaning |
|--------|------|---------|
| `#N` or `N` | Immediate | Literal value N (0-255) |
| `@N` | Direct | Value stored at memory address N |
| `*N` | Indirect | Read address from memory[N], then value there |

Numbers are decimal by default. Use 0x prefix for hex (e.g., 0xFF).

## Instruction Set (each instruction = 4 bytes)

### Data Movement
| Instruction | Effect |
|-------------|--------|
| `MOV src dst` | Copy src value to dst address |
| `MOV src dst count` | Fill count bytes at dst with src value (if src is immediate), or copy count consecutive bytes from src to dst. |

### Arithmetic (all mod 256)
| Instruction | Effect |
|-------------|--------|
| `ADD a b dst` | memory[dst] = (a + b) mod 256 |
| `SUB a b dst` | memory[dst] = (a - b) mod 256 |
| `MUL a b dst` | memory[dst] = (a * b) mod 256 |
| `DIV a b dst` | memory[dst] = a / b (integer, 255 if b=0) |
| `MOD a b dst` | memory[dst] = a % b (255 if b=0) |

### Control Flow
| Instruction | Effect |
|-------------|--------|
| `JMP #N` | Jump by N instructions relative to current (0=no-op, 1=skip next) |
| `JMP @N` | Jump to absolute byte address stored in memory[N] |
| `JEQ a b #N` | If a == b, jump by N instructions relative |
| `JEQ a b @N` | If a == b, jump to absolute address in memory[N] |
| `JNE a b #N/@N` | Same but if a != b |
| `JGR a b #N/@N` | Same but if a > b |

**Backward jumps**: Values are 0-255 (no negatives). For backward jumps, store the target byte address in a memory cell and use @cell. Instruction N is at byte address N*4.

### Graphics
| Instruction | Effect |
|-------------|--------|
| `PIX addr color` | Set pixel at linear address to color (low 4 bits) |

### Other
| Instruction | Effect |
|-------------|--------|
| `FLP a b` | Swap values at addresses a and b |
| `THR addr` | Spawn a new thread starting at the given address |

## Memory Layout
```
Row 0:  addr 0-15    (y=0)
Row 1:  addr 16-31   (y=1)
...
Row 15: addr 240-255 (y=15)

Pixel at (x,y) = address y*16 + x
```

## Key Constraints
- Max 64 instructions (4 bytes each = 256 bytes total)
- Code occupies the first N*4 addresses. Those pixels will show code bytes.
- Use high addresses (@248-@255) for variables/counters.
- Programs loop implicitly after the last instruction.

# RESPONSE FORMAT

Reason through the problem in <think> tags, then provide your solution in <answer> tags.
The code inside <answer> will be executed directly.
\end{Verbatim}
\end{tcolorbox}
\vspace{4pt}